\documentclass[twoside,11pt,preprint]{article}

\PassOptionsToPackage{hyphens}{url}

\usepackage{dmlr2e_preprint}

\usepackage{array}
\usepackage{booktabs}
\usepackage{graphicx}
\usepackage{multirow}
\usepackage[caption=false]{subfig}
\usepackage{tabularx}
\usepackage{url}
\usepackage{xcolor}

\newcommand{\monster}{\textsc{Monster}}

\newcommand{\zo}{\mbox{0--\!1}~loss}

\newcommand{\fcn}{\textsc{FCN}}
\newcommand{\hinception}{\textsc{H-InceptionTime}}
\newcommand{\tempcnn}{\textsc{TempCNN}}
\newcommand{\rocket}{\textsc{Rocket}}
\newcommand{\quant}{\textsc{Quant}}
\newcommand{\hydra}{\textsc{Hydra}}

\usepackage{lastpage}
\dmlrheading{25}{2025}{1-\pageref{LastPage}}{02/25; Revised XX/XX}{XX/XX}{25-0000}{Dempster et al.}
\def\openreview{\null}

\ShortHeadings{\textit{MONSTER}}{Dempster et al.}
\firstpageno{1}

\begin{document}

\vspace*{5ex}

\title{\textit{MONSTER}\\Monash Scalable Time Series Evaluation Repository}

\author{%
    \name Angus Dempster\footnotemark[1] \email angus.dempster@monash.edu \\
    \name Navid Mohammadi Foumani\footnotemark[1] \\ 
    \name Chang Wei Tan \\
    \name Lynn Miller\footnotemark[1] \\
    \name Amish Mishra\footnotemark[1] \\
    \name Mahsa Salehi\footnotemark[1] \\
    \name Charlotte Pelletier\footnotemark[2] \\
    \name Daniel F. Schmidt\footnotemark[1] \\
    \name Geoffrey I. Webb\footnotemark[1] \\
    \addr \footnotemark[1] Monash University, Melbourne, Australia \\
    \addr \footnotemark[2] Université Bretagne Sud, IRISA, Vannes, France
}

\editor{TBA}

\maketitle


\begin{abstract}
    We introduce {\monster}---the \textbf{MON}ash \textbf{S}calable \textbf{T}ime Series \textbf{E}valuation \textbf{R}epository---a collection of large datasets for time series classification. The field of time series classification has benefitted from common benchmarks set by the UCR and UEA time series classification repositories. However, the datasets in these benchmarks are small, with median sizes of 217 and 255 examples, respectively. In consequence they favour a narrow subspace of models that are optimised to achieve low classification error on a wide variety of smaller datasets, that is, models that minimise variance, and give little weight to computational issues such as scalability. Our hope is to diversify the field by introducing benchmarks using larger datasets. We~believe that there is enormous potential for new progress in the field by engaging with the theoretical and practical challenges of learning effectively from larger quantities of data.
\end{abstract}



\section{Introduction} \label{sec-introduction}

`State of the art' in time series classification has become synonymous with state of the art on the datasets in the UCR and UEA archives \citep{UEA-2018,UCR-2019,bagnall_etal_2017,middlehurst_etal_2024,ruiz_etal_2021}. However, most of these datasets---at least, most of those that are commonly used for evaluation---are small: median training set size for the set of 142 canonical univariate time series datasets is just 217 examples. The preeminence of the datasets in the UCR and UEA archives as a basis for benchmarking means that the field has become constrained by a narrow focus on smaller datasets and models which achieve low {\zo} (classification error) on a diversity of smaller datasets.

Empirical machine learning research relies heavily on benchmarking in one form or another \citep{liao_etal_2021}. Benchmark datasets provide the data necessary for training and evaluating machine learning models. Certain datasets and benchmarks have become foundational to machine learning generally \citep{paullada_etal_2021}. There is little doubt that the datasets in the UCR and UEA archives are as integral to the field of time series classification as are, for example, the MNIST, CIFAR, and ImageNet datasets to the field of image classification.

`[T]he ways in which we collect, construct, and share these datasets inform the kinds of problems the field pursues and the methods explored in algorithm development' \citep{paullada_etal_2021}. We might call this the `dataset lottery' or `benchmark lottery'---after the `hardware lottery'---i.e., to paraphrase \citet{hooker_2021}, when a method or set of methods `win' (predominate) because of their compatibility with existing benchmarks.

A benchmark should serve as a proxy for a broader task (e.g., image classification, or time series classification). A given benchmark is only meaningful to the extent that performance on that benchmark reflects performance on a broader task, and performance on that benchmark generalises to real-world problems \citep{liao_etal_2021}.

In the context of time series classification, current benchmarks favour models that have been optimised to achieve low classification error ({\zo}) on a diversity of smaller datasets, i.e., low-variance (high-bias) methods: see Section \ref{sec-background}. Datasets currently used for benchmarking do not reflect either the theoretical or practical challenges of learning from large-scale real-world data.

This poses the risk that current benchmarks are unrepresentative of the broader task of time series classification, and that models considered state of the art on these benchmarks may not generalise to---and therefore may have diminishing relevance for---real-world time series classification problems, especially those involving larger quantities of data. This also suggests that research in time series classification has only so far explored a relatively narrow subset of ideas \citep[see][]{hooker_2021}.

We present {\monster}---the \textbf{Mon}ash \textbf{S}calable \textbf{T}ime Series \textbf{E}valuation \textbf{R}epository---a collection of large univariate and multivariate datasets for time series classification. Our aim is to complement the existing datasets in the UCR and UEA archives, while encouraging the field to diversify to include significantly larger datasets. We hope that, with the introduction of {\monster}, benchmarking in the field better represents the broader task of time series classification, and has increased relevance for real-world time series classification problems. We hope to inspire the field to engage with the challenges of learning from large quantities of data. We believe that there is enormous potential for new progress in the field.

The rest of this paper is structured as follows. Section 2 expands on relevant background material. Section 3 provides further details of the {\monster} datasets. Section 4 provides preliminary baseline results for selected methods.


\section{Background} \label{sec-background}

\subsection{Bias--Variance Tradeoff}

A benchmark should reflect a broader task, and performance on a given benchmark should generalise to real-world problems. In this context, the UCR and UEA archives share one key limitation. Whereas historically, in the field of computer vision, different methods have generally been evaluated on a relatively small number of large datasets (e.g., ImageNet), in the field of time series classification, different methods are almost always evaluated on a relatively large number of small datasets, i.e., the datasets in the UCR and UEA archives.

\begin{figure}
    \centering%
    \includegraphics[width=0.5\linewidth]{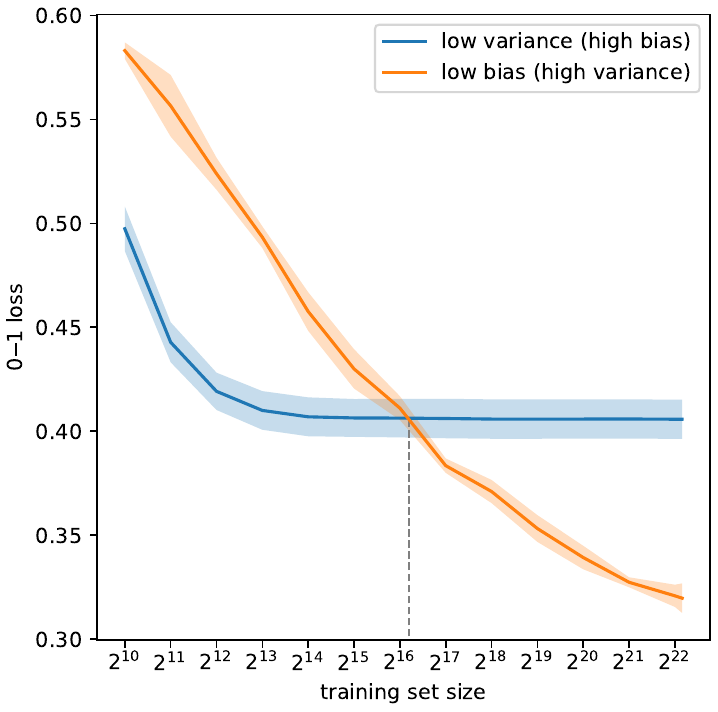}%
    \caption{Learning curves for a low variance model vs a low bias model on \textit{S2Agri-10pc-17}.}%
    \label{fig-bv}
\end{figure}

Variance can be expected to be large when training sets are small and to decrease as training set size increases. As a result, methods that effectively minimise variance will often achieve lower classification error on smaller datasets, while methods that minimise bias will often achieve lower classification error on larger datasets \citep{brain1999effect}. This is illustrated in Figure \ref{fig-bv}, which shows learning curves for two models: a low variance model (a single-layer CNN with random kernels) vs a low bias model (a conventional two-layer CNN) on the \textit{S2Agri-10pc-17} dataset (further details are provided in Appendix \ref{sec-appendix-bv}). Figure~\ref{fig-bv} shows that the low-variance model achieves lower {\zo} on smaller quantities of data, whereas the low-bias model achieves lower {\zo} on larger quantities of data. (Note that the appearance of learning curves such as these is potentially confounded by multiple factors, including how well the biases of the respective systems match the characteristics of the learning task.)

We should not expect the same methods to achieve the lowest {\zo} on both smaller datasets and larger datasets, as these demand different learning characteristics. (The issue of dataset size is not just limited to the quantity of training data: small quantities of test data can mask large differences in real-world classification error: \citeauthor{liao_etal_2021}, \citeyear{liao_etal_2021}.)

As such, the methods currently considered state of the art in terms of accuracy on the datasets in the UCR and UEA archives are, by definition, likely dominated by methods optimised for smaller datasets or, in other words, methods that minimise variance (high bias, low variance methods). We see strategies for minimising variance in all or almost all state of the art methods for time series classification. Variance can be minimised via ensembling (e.g., InceptionTime \citep{Fawaz2019a}, the HIVE-COTE models \citep{middlehurst_etal_2021}, Proximity Forest \citep{lucas_etal_2019}, and models such as DrCIF \citep{middlehurst_etal_2021}  using ensembles of decision trees), explicit regularisation (e.g., methods using a ridge classifier such as RDST \citep{guillaume_etal_2022}, Weasel 2.0 \citep{schaefer_etal_2023}), and/or overparameterisation (taking advantage of double descent, e.g., the {\rocket} `family' of methods \citep{dempster_etal_2020,dempster_etal_2021,dempster_etal_2023}, and other methods making use of a large feature space in combination with a ridge regression classifier or other linear model, as well as large neural network models), or some combination of these approaches.

\subsection{The `Bitter Lesson'}

It is not a coincidence that, with some exceptions, deep learning methods have had a relatively muted impact on the field. Models such as large deep neural networks are low bias models, and require significant quantities of training data in order to achieve competitive accuracy compared to less complex models. There has been a significant amount of work applying deep learning methods in the field of time series classification~\citep{navid2024}. However, despite this, and despite the fact that some neural network models such as InceptionTime \citep{Fawaz2019a} are among the most accurate models, on average, on the datasets in the UCR and UEA archives, in large part deep learning methods have not had the kind of impact that they have had in other domains such as image classification or natural language processing.

Arguably, time series classification has not yet had its `ImageNet moment', simply because in almost all existing work the quantity of training data has been insufficient to allow for training low bias models such as large convolutional neural networks or transformer architectures effectively. (A not insubstantial amount of work involving deep learning in the context of time series is also problematic, e.g., involving directly or indirectly optimising test loss: \citet{middlehurst_etal_2024}.)

It is not clear yet whether the `bitter lesson'---`the only thing that matters in the long run is the leveraging of computation' \citep{sutton_2019}---has yet been learned in the field of time series classification. The apparent diversity of methods considered state of the art may reflect a diversity of inductive biases that are effective for extracting information from low quantities of data, but that actually limit the ability to learn effectively from large quantities of data.

There is also the potential issue of overfitting a benchmark itself, although this is of less immediate concern due to the recent additional of new datasets to the UCR archive \citep{middlehurst_etal_2024}. Accordingly, as well as being larger, the {\monster} datasets also represent new datasets or, in other words, a new `out of sample' collection of datasets on which to evaluate existing methods.

\subsection{`No Free Lunch'}

Evaluation on a large set of heterogeneous datasets has led to another difference (in contrast to, e.g., computer vision), namely, that in the field of time series classification, performance is typically measured in terms of accuracy over all of the datasets in the UCR and/or UEA archives. This kind of average performance represents an average over a large set of highly heterogeneous input time series datasets.

This favours, without necessarily any good reason, methods that perform well (achieve low classification error) \textit{on average}, while not necessarily performing well on any particular subset of datasets or tasks.

The `no free lunch' theorem suggests that, as the number of datasets included in the evaluation grows, the performance of all methods should converge \textit{on average}, i.e., no one method will perform better than any other on all datasets \citep{wolpert_and_macready_1997}. In the real world, this kind of average performance is potentially of limited practical value. For example, given a problem involving the classification of EEG data, we would rather use a method demonstrated to have good classification performance on benchmark EEG data, rather than a method that has low \textit{average} classification error across both EEG data and data from one or more other domains.

In other words, current research likely unjustifiably favours methods that not only minimise variance, but that achieve low {\zo} \textit{on average}, with potentially limited relevance to any specific real-world application.

In many cases it makes sense for a model or architecture to be specialised to a particular domain. For example, TempCNN uses short convolutional kernels, ideal for the short time series typical of Earth observation data, but which are not effective for capturing temporal relationships in long time series, e.g., those common in audio tasks. The lack of pooling layers allows TempCNN to locate temporal features important for tasks such as crop detection, but lacks the ability to detect scale-invariant features important in some other tasks~\citep{Pelletier2019}. In contrast, ConvTran uses channel-wise convolutional kernels and attention to capture both relationships between channels and long-range temporal relationships, especially effective for EEG data \citep{ConvTran}, but which have potentially limited relevance to univariate and/or shorter time series.

\subsection{Other Selection Pressures and the `Hardware Lottery'}

For the most part, the field has not been forced to contend with the practical challenges involved with learning from larger quantities of data. Just as smaller datasets favour methods that effectively minimise variance, different kinds of selection pressures exist in the context of larger datasets.

In particular, larger datasets select for methods that are computationally suited to large datasets, and can make effective use of existing computational resources, i.e., the `hardware lottery' (Hooker, 2021). Methods with high computational and/or memory requirements quickly become impractical. Even for more efficient methods, training on large quantities of data presents significant practical challenges.

\subsection{Opportunities}

The need for expanding benchmarking in the field to include larger datasets has been recognised for some time. Dau et al (2019) stated: `[p]erhaps a specialist archive of massive time series can be made available for the community in a different repository' (p 1295).

{\monster} represents an opportunity for the field to diversify to include large datasets, to engage with the challenges of learning from larger datasets, to better reflect the broader task of time series classification, and to improve relevance for real-world time series classification problems. We believe that there is enormous opportunity for new progress in the field.

Further, we make the following predictions in relation to the ways in which larger datasets might change the field of time series classification, which may or may not be borne out in practice in the long run:%
\begin{itemize}%
\item Only a subset of existing methods will be practical, i.e., those which can take advantage of current hardware to train efficiently.%
\item The methods which achieve the lowest {\zo} on larger datasets will differ from the methods which achieve the lowest {\zo} on smaller datasets.%
\item Average performance (e.g., average {\zo}) will become less relevant than performance within meaningful subsets of tasks (e.g., classification of EEG data, vs classification of satellite image time series data).%
\end{itemize}


\section{The MONSTER Datasets} \label{sec-monster}


The initial release of the {\monster} benchmark includes 29 univariate and multivariate datasets with between $10{,}299$ and $59{,}268{,}823$ time series. Table \ref{table:Datasets} provides an overview of the datasets. (We consider this as an initial release, and we aim to continue to add datasets to the benchmark.)  The datasets are available via HuggingFace: \url{https://huggingface.co/monster-monash}). Relevant code is available at: \url{https://github.com/Navidfoumani/monster}. We provide the datasets in \texttt{.npy} format to allow for ease of use with Python and straightforward memory mapping. (We also provide the datasets in legacy \texttt{.csv} format.)  All datasets are under creative commons licenses or in the public domain, or we otherwise have been given permission to include the dataset in this collection. All datasets are already publicly available in some form.

We have processed the original time series into a common format (\texttt{.npy} and \texttt{.csv}). The steps required to process each dataset were different and included, for example, extracting and labelling individual time series from broader time series data, interpolating irregularly sampled data, and resampling data where the original data was recorded at different sampling rates. We have endeavoured to lower the `barrier of entry' as much as possible while keeping the original data intact to the greatest extent possible. Further details for each of the datasets are set out below.

\begin{figure}[h]%
    \centering%
    \includegraphics[width=0.75\linewidth]{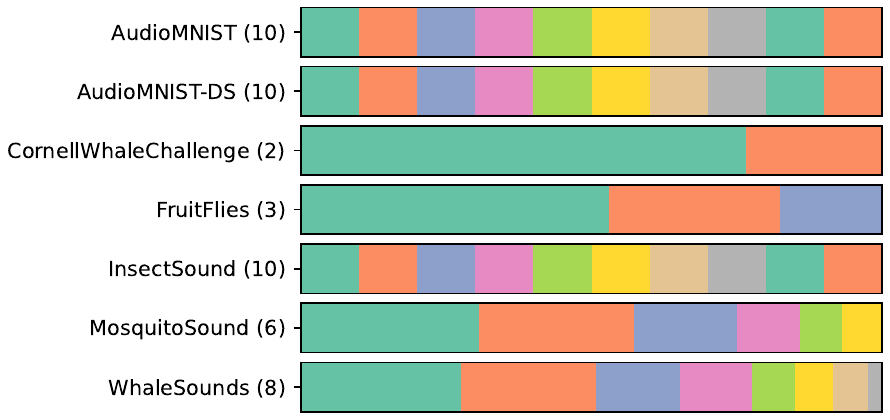}%
    \caption{Class distributions for the audio datasets.}%
    \label{fig-class-dist-audio}
\end{figure}

\begin{figure}[h]%
    \centering%
    \includegraphics[width=0.75\linewidth]{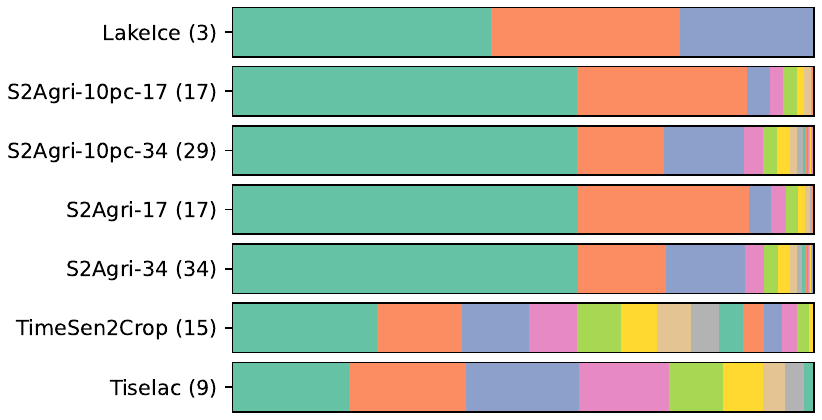}%
    \caption{Class distributions for the satellite datasets.}%
    \label{fig-class-dist-satellite}
\end{figure}

\begin{figure}[h]%
    \centering%
    \includegraphics[width=0.75\linewidth]{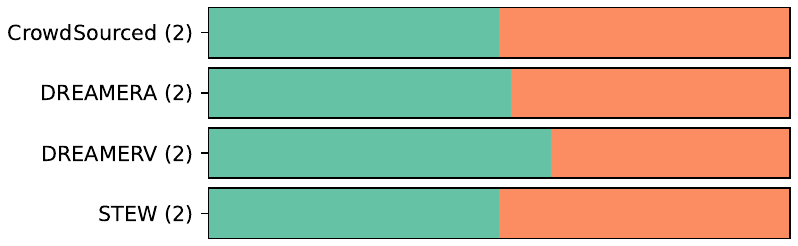}%
    \caption{Class distributions for the EEG datasets.}%
    \label{fig-class-dist-eeg}
\end{figure}

\begin{figure}[h]%
    \centering%
    \includegraphics[width=0.75\linewidth]{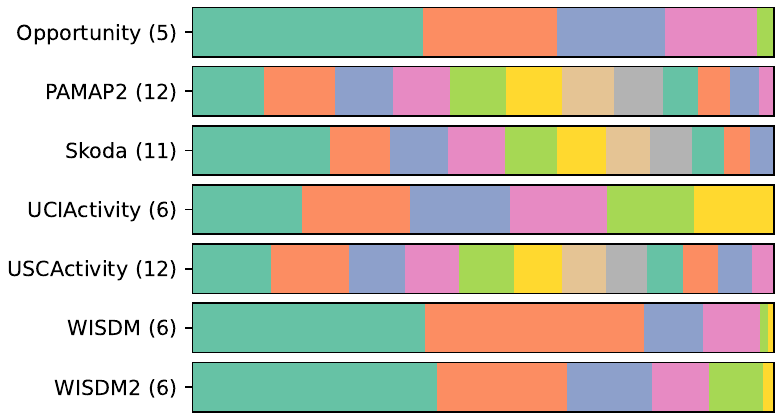}%
    \caption{Class distributions for the HAR datasets.}%
    \label{fig-class-dist-har}
\end{figure}

\begin{figure}[h]%
    \centering%
    \includegraphics[width=0.75\linewidth]{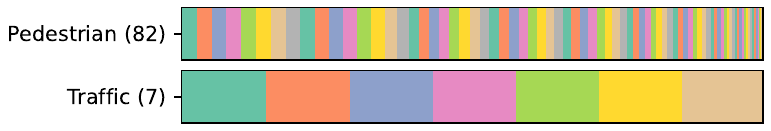}%
    \caption{Class distributions for the count datasets.}%
    \label{fig-class-dist-count}
\end{figure}

\begin{figure}[h]%
    \centering%
    \includegraphics[width=0.75\linewidth]{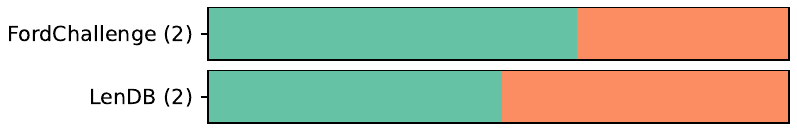}%
    \caption{Class distributions for the uncategorised datasets.}%
    \label{fig-class-dist-other}
\end{figure}

Each dataset is provided with a set of indices for 5-fold cross-validation, allowing for direct comparison between benchmark results. For some datasets, these simply represent stratified random cross-validation folds. For other datasets, the cross-validation folds have been generated taking into account important metadata, e.g., different experimental subjects (for EEG data), or different geographic locations (for satellite image time series data). We have assigned the datasets to one of six categories (audio, satellite, EEG, HAR, count, and other). The distribution of classes for the datasets in each category is shown in Figures \ref{fig-class-dist-audio}, \ref{fig-class-dist-satellite}, \ref{fig-class-dist-eeg}, \ref{fig-class-dist-har}, \ref{fig-class-dist-count}, and \ref{fig-class-dist-other}, below (in each figure, the number in brackets corresponds to the number of classes).

\begin{table}
    \centering
    \begin{tabular}{lrrrr}
        \toprule
        \textbf{Dataset} & \textbf{Instances} & \textbf{Length} & \textbf{Channels} & \textbf{Classes} \\
        \midrule
        \multicolumn{5}{c}{Audio} \\
        \midrule
        AudioMNIST & 30{,}000 & 47{,}998 & 1 & 10 \\
        AudioMNIST-DS & 30{,}000 & 4{,}000 & 1 & 10 \\
        CornellWhaleChallenge & 30{,}000 & 4{,}000 & 1 & 2 \\
        FruitFlies & 34{,}518 & 5{,}000 & 1 & 3 \\
        InsectSound & 50{,}000 & 600 & 1 & 10 \\
        MosquitoSound & 279{,}566 & 3{,}750 & 1 & 6 \\
        WhaleSounds & 105{,}163 & 2{,}500 & 1 & 8 \\
        \midrule
        \multicolumn{5}{c}{Satellite Image Time Series} \\
        \midrule
        LakeIce & 129{,}280 & 161 & 1 & 3 \\
        S2Agri & 59{,}268{,}823 & 24 & 10 & 17 / 34 \\
        S2Agri-10pc & 5{,}850{,}881 & 24 & 10 & 17 / 29 \\
        TimeSen2Crop & 1{,}135{,}511 & 365 & 9 & 16 \\
        Tiselac & 99{,}687 & 23 & 10 & 9 \\
        \midrule
        \multicolumn{5}{c}{EEG} \\
        \midrule
        CrowdSourced & 12,289 & 256 & 14 & 2 \\
        DreamerA & 170,246 & 256 & 14 & 2 \\
        DreamerV & 170,246 & 256 & 14 & 2 \\
        STEW & 28,512 & 256 & 14 & 2 \\
        \midrule
        \multicolumn{5}{c}{Human Activity Recognition} \\
        \midrule
        Opportunity & 17,386 & 100 & 113 & 5 \\
        PAMAP2 & 38,856 & 100 & 52 & 12 \\
        Skoda & 14,117 & 100 & 60 & 11 \\
        UCIActivity & 10{,}299 & 128 & 9 & 6 \\
        USCActivity & 56,228 & 100 & 6 & 12 \\
        WISDM & 17,166 & 100 & 3 & 6 \\
        WISDM2 & 149,034 & 100 & 3 & 6 \\
        \midrule
        \multicolumn{5}{c}{Counts} \\
        \midrule
        Pedestrian & 189{,}621 & 24 & 1 & 82 \\
        Traffic & 1{,}460{,}968 & 24 & 1 & 7 \\
        \midrule
        \multicolumn{5}{c}{Other} \\
        \midrule
        FordChallenge & 36,257 & 40 & 30 & 2 \\
        LenDB & 1{,}244{,}942 & 540 & 3 & 2 \\
        \bottomrule
    \end{tabular}
    \caption{Summary of {\monster} datasets.}
    \label{table:Datasets}
\end{table}

\subsection{Audio}

\subsubsection{AudioMNIST and AudioMNIST-DS}

\textbf{\textit{AudioMNIST}} consists of audio recordings of 60 different speakers saying the digits 0 to 9, with 50 recordings per digit per speaker \citep{data_audiomnist_2024,becker_etal_2024}. The processed dataset contains $30{,}000$ (univariate) time series, each of length $47{,}998$ (approximately 1 second of data sampled at 44khz), with ten classes representing the digits 0 to 9. This version of the dataset has been split into cross-validation folds based on speaker (i.e., such that recordings for a given speaker do not appear in both the training and validation sets). \textbf{\textit{AudioMNIST-DS}} is a variant of the same dataset downsampled to a length of $4{,}000$.

\subsubsection{CornellWhaleChallenge}

\textbf{\textit{CornellWhaleChallenge}} consists of hydrophone recordings \citep{data_rightwhales_2013}. The processed dataset consists of $30{,}000$ (univariate) time series, each of length $4{,}000$. The task is to distinguish right whale calls from other noises. (An abridged version of this dataset is included in the broader UCR archive.)  This version of the dataset has been divided into stratified random cross-validation folds.

\subsubsection{FruitFlies}

\textbf{\textit{FruitFlies}}, taken from the broader UCR archive, consistst of $34{,}518$ (univariate) time series, each of length $5{,}000$, representing acoustic recordings of wingbeats for three species of fruit fly \citep{data_fruitflies_2016,flynn_2022}. The task is to identify the species of fly based on the recordings. This version of the dataset has been split into stratified random cross-validation folds.

\subsubsection{InsectSound}

\textbf{\textit{InsectSound}}, taken from the broader UCR archive, consists of $50{,}000$ (univariate) time series, each of length $600$, representing recordings of wingbeats for six species of insects, with 2 different genders for 4 of the 6 species \citep{chen_etal_2014,data_insectsound_2014}. This version of the dataset has been split into stratified random cross-validation folds.

\subsubsection{MosquitoSound}

\textbf{\textit{MosquitoSound}}, taken from the broader UCR archive, consists of $279{,}566$ (univariate) time series, each of length $3{,}750$, representing recordings of wingbeats for six different species of mosquito \citep{fanioudakis_etal_2018,data_mosquitosound_2018}. The task is to identify the species of mosquito based on the recordings. This version of the dataset has been split into stratified random cross-validation folds.

\subsubsection{WhaleSounds}

\textbf{\textit{WhaleSounds}} consists of underwater acoustic recordings around Antarctica, manually annotated for seven different types of whale calls \citep{data_whalesounds_2020,miller_etal_2021}. The dataset has been processed to extract the annotated whale calls from the original recordings. The processed dataset contains $105{,}163$ (univariate) time series, each of length $2{,}500$, with eight classes representing the seven types of whale call plus a class for unidentified sounds. This version of the dataset has been split into stratified random cross-validation folds.

\subsection{Satellite Time Series}

\subsubsection{LakeIce}

\textbf{\textit{LakeIce}} consists of pixel-level backscatter (reflection) values from satellite images of Yukon, Canada \citep{data_lakeice_2022,shaposhnikova_eal_2023}. The time series are extracted over three decades from ERS-1/2, Radarsat, and Sentinel-1 synthetic aperture radar satellites. The processed dataset contains $129{,}280$ (univariate) time series each of length $161$, representing 6 months of data (October to March), with three classes representing bedfast ice, floating ice, and land. This version of the dataset has been split into stratified random cross-validation folds.

\subsubsection{S2Agri}

\begin{figure}
\centering
     \includegraphics[trim=0cm 0cm 0cm 0cm, width=0.48\columnwidth]{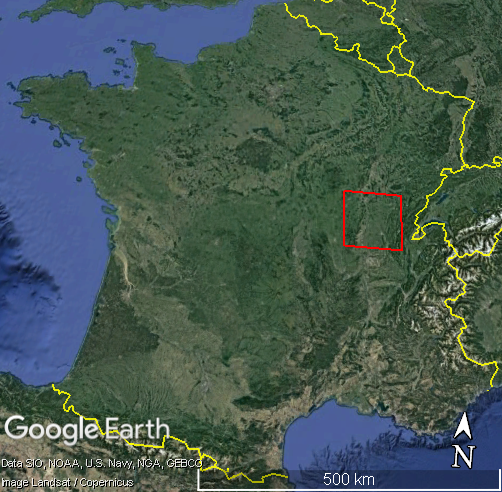}
     \caption{Map of France showing the location of the Sentinel-2 tile used in the S2Agri dataset.}
     \label{fig:S2agri}
\end{figure}

\textbf{\textit{S2Agri}} is a land cover classification dataset and contains a single tile of Sentinel-2 data (T31TFM), which covers a $12{,}100$ km2 area in France: see Figure \ref{fig:S2agri}~\citep{garnot_etal_2020,data_s2agri_2022}. Ten spectral bands are used, and these are provided at 10m resolution. The dataset contains time series of length $24$, observed between January and October 2017. The area has a wide range of crop types and terrain conditions.

The original S2Agri dataset is designed for parcel-based processing and contains data for $191{,}703$ land parcels, with data for each parcel provided in a separate file. We have reorganised the data for pixel-based processing, leading to a dataset containing $59{,}268{,}823$ pixels. Two sets of land cover classification labels are provided, one with $19$ classes and the other with $44$ classes. However, some of the 44-classes are only represented by one land parcel. We have removed the pixels in these land parcels from the dataset. This means there are only $17$ and $34$ classes respectively that are represented in the final dataset. The class label of each parcel comes from the French Land Parcel Identification System. The dataset is unbalanced: the largest four of the 19-class labels account for $90\%$ of the parcels.

We thus provide two versions of the S2Agri dataset, \textbf{\textit{S2Agri-17}}, which uses the $17$ class labels and \textbf{\textit{S2Agri-34}}, which uses the $34$ class labels. Additionally, we have created smaller versions of the datasets consisting of data for a randomly selected $10\%$ of the land parcels, each containing $5{,}850{,}881$ pixels. These are \textbf{\textit{S2Agri-10pc-17}} and \textbf{\textit{S2Agri-10pc-34}} for the $17$-class and $34$-class labels, respectively.

To create the folds used for cross-validation, we split the data based on the original land parcels, thus ensuring that all pixels in a land parcel are allocated to the same fold. Splits are stratified by class labels to ensure an even representation of the classes across the folds.

\subsubsection{TimeSen2Crop}

\begin{figure} 
    \centering
    \subfloat[Map of Austria showing Sentinel-2 Tiles]{%
        \includegraphics[width=0.52\textwidth]{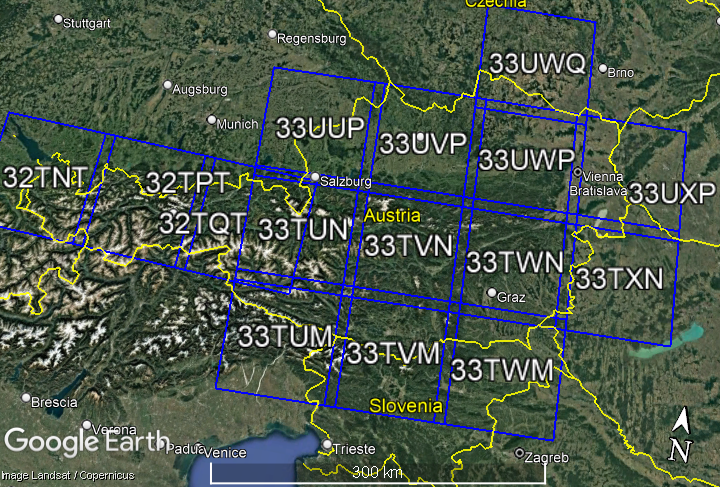}%
        \label{fig:timesen2crop-a}%
        }%
    \hfill%
    \subfloat[Class counts by Sentinel-2 Tile]{%
        \includegraphics[width=0.46\textwidth]{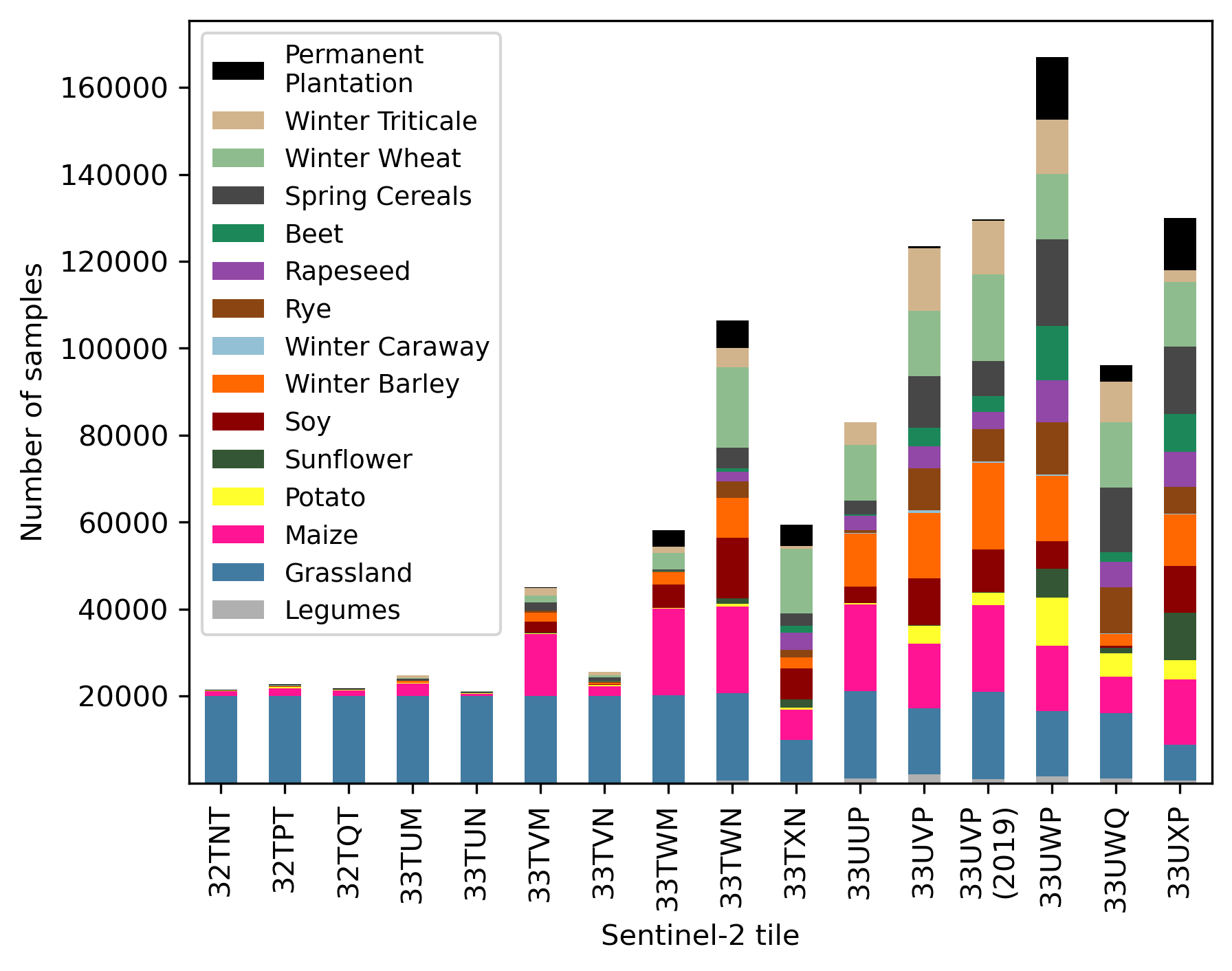}%
        \label{fig:timesen2crop-b}%
        }%
    \caption{Location of Sentinel-2 tiles and class counts for the TimeSen2Crop dataset}
\end{figure}

\textbf{\textit{TimeSen2Crop}} consists of pixel-level Sentinel-2 data at a 10m resolution, extracted from the 15 Sentinel-2 tiles that cover Austria: see Figure \ref{fig:timesen2crop-a}~\citep{Weikmann2021}. The dataset contains 16 classes representing different land cover types. The original data contains all Sentinel 2 images covering Austria acquired between September 2017 and August 2018 plus images for one tile acquired between September 2018 and August 2019. As the tiles are from different Sentinel-2 tracks and have been processed to remove images with cloud cover greater than 80\%, the image acquisition dates for each tile differ and are irregular. This version of the dataset has been processed to interpolate each pixel to a daily time series representing one year of data (thus each pixel has a time series length of 365) and removing the ``other crops'' class. The processed dataset contains $1{,}135{,}511$ multivariate time series, each with 9 channels (representing 9 spectral bands) and 15 classes. Classes are unbalanced and unevenly distributed across the Sentinel-2 tiles: see Figure \ref{fig:timesen2crop-b}. The dataset has been split into cross-validation folds based on geographic location by Sentinel-2 tile (i.e., such that, for each fold, time series from a given location appear in either the training set or test set but not both).

\subsubsection{TiSeLaC}

\begin{figure} 
    \centering
    \subfloat[Map of Reunion Island and fold data distribution]{%
        \includegraphics[width=0.48\textwidth]{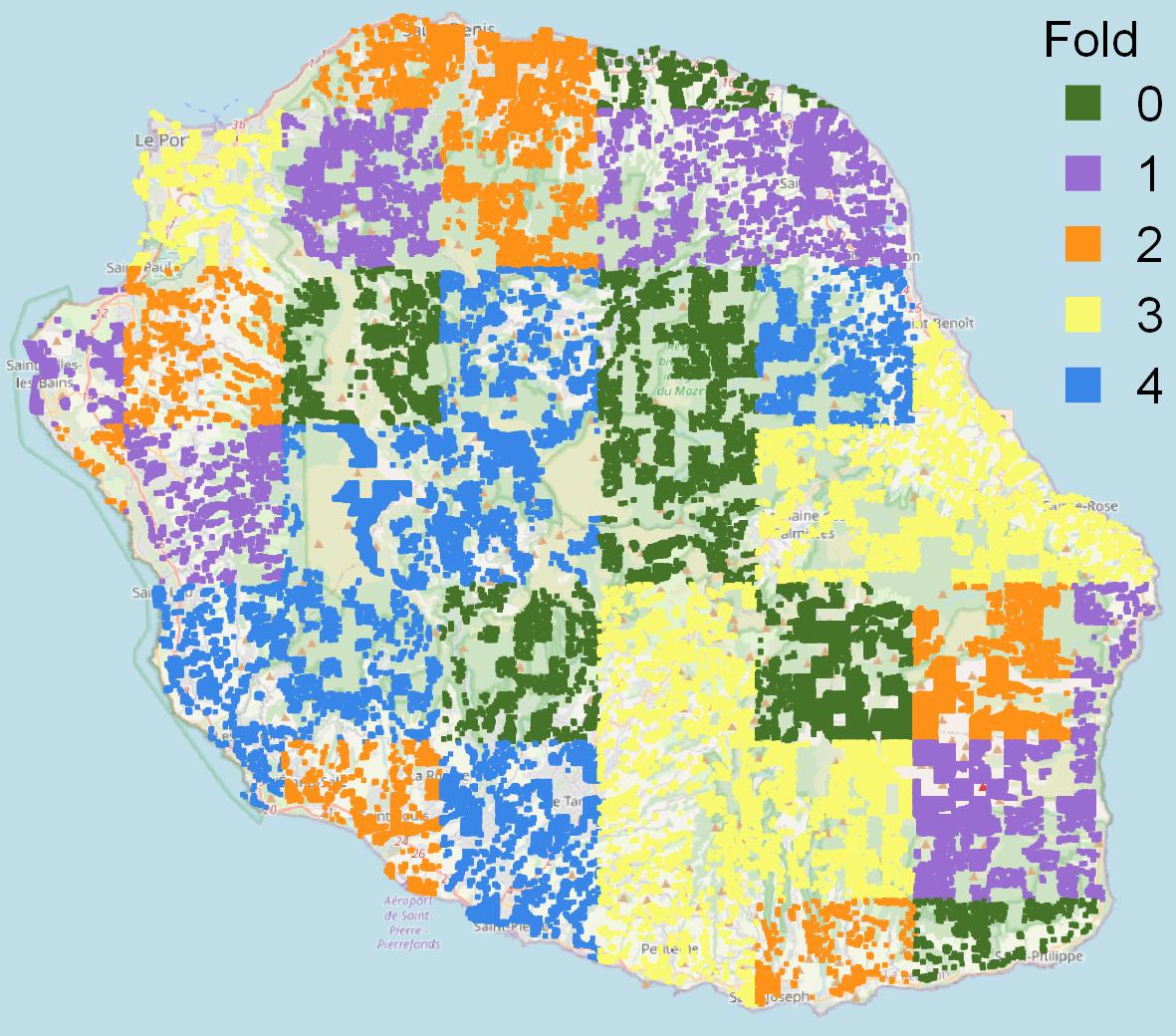}%
        \label{fig:tiselac-a}%
        }%
    \hfill%
    \subfloat[Label counts by fold]{%
        \includegraphics[width=0.50\textwidth]{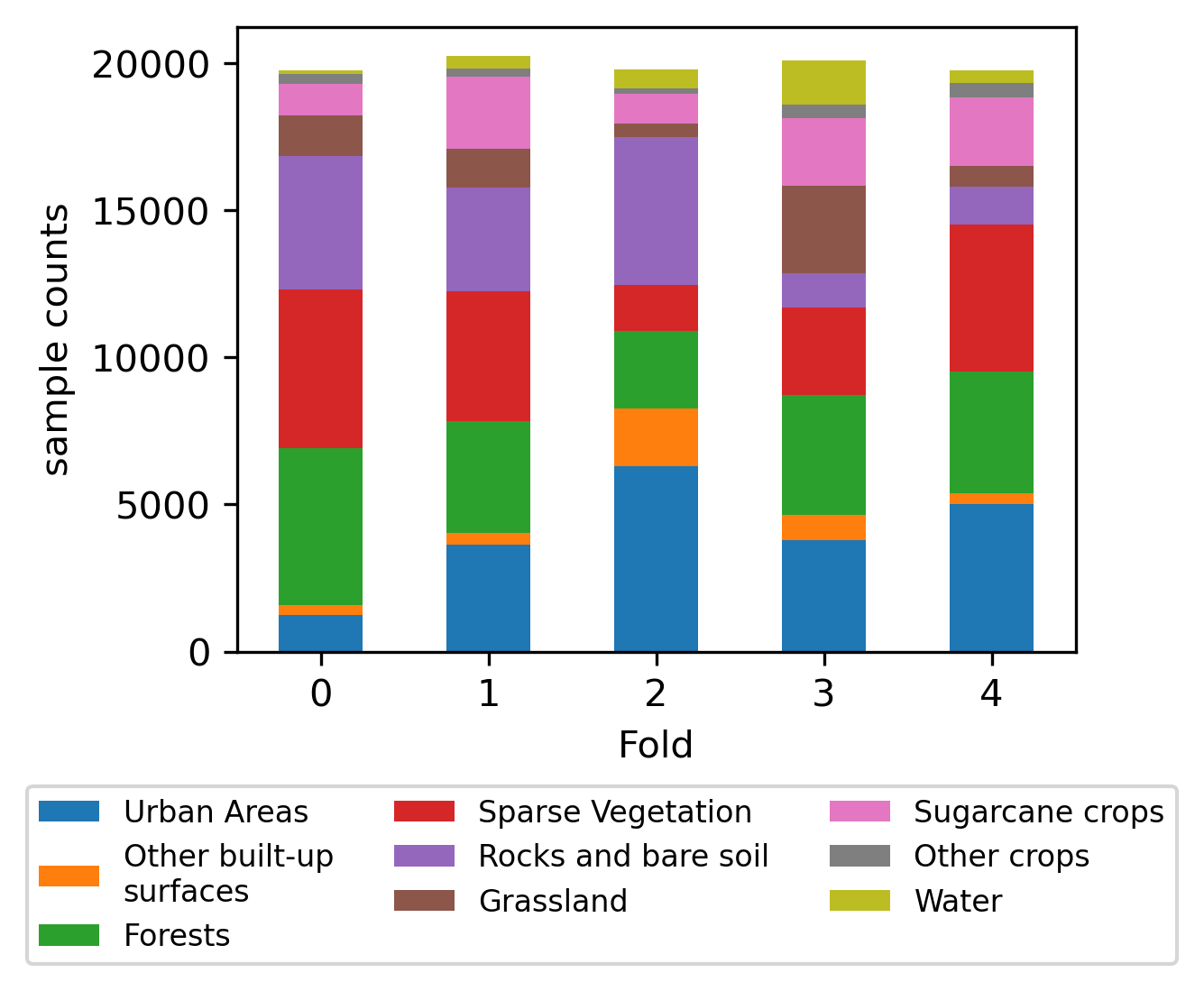}%
        \label{fig:tiselac-b}%
        }%
    \caption{Map of Reunion Island and label counts by fold for the Tiselac dataset. Note (a); Map from Open Street Map, sample data pixels are not to scale.}
\end{figure}

\textbf{\textit{TiSeLaC}} (Time Series Land Cover Classification) was created for the time series land cover classification challenge held in conjunction with the 2017 European Conference on Machine Learning \& Principles and Practice of Knowledge Discovery in Databases~\citep{TiSeLaC2017}. It was generated from a time series of 23 Landsat 8 images of Reunion Island (Figure \ref{fig:tiselac-a}), acquired in 2014. At the 30m spatial resolution of Landsat 8 images, Reunion Island is covered by $2866 \times 2633$ pixels, however only 99,687 of these pixels are included in the dataset.
Class labels were obtained from the 2012 Corine Land Cover (CLC) map and the 2014 farmers' graphical land parcel registration (Régistre Parcellaire Graphique - RPG) and the nine most significant classes have been included in the dataset. The number of pixels in each class is capped at 20,000. The data was obtained from the winning entry's GitHub repository~\citep{DiMauro2017}, which includes the spatial coordinates for each pixel.

We provide training and testing splits designed to give spatial separation between the splits, which should lead to realistic estimations of the generalisation capability of trained models. We first divided the original pixel grid into a coarse grid, with each grid cell sized at $300 \times 300$ pixels, then computed the number of dataset pixels in each cell (the cell size). These cells are processed in descending order of size, and allocated to the fold with the fewest pixels. The resulting spatial distribution of the folds is shown in Figure \ref{fig:tiselac-a} and the distribution of classes across the folds is shown in Figure \ref{fig:tiselac-b}.

\subsection{EEG}

\subsubsection{CrowdSourced}

\textbf{\textit{CrowdsSourced}} consists of EEG data collected as part of a study investigating brain activity during a resting state task, which included two conditions: \textit{eyes open} and \textit{eyes closed}, each lasting 2 minutes. The dataset contains EEG recordings from 60 participants, but only 13 successfully completed both conditions. The recordings were captured using 14-channel EEG headsets—specifically the \textit{Emotiv EPOC+}, \textit{EPOC X}, and \textit{EPOC} devices. These devices provide high-quality, wireless brainwave data that is ideal for analyzing resting-state brain activity~\citep{crowdsourced}.

The data was initially recorded at a high frequency of 2048 Hz and later downsampled to 128 Hz for processing. To segment the data for analysis, we used a 2-second window (equivalent to 256 time steps) with a 32 time-step stride to capture the dynamics of brain activity while maintaining a manageable data size. The raw EEG data for the 13 participants, along with preprocessing steps, analysis scripts, and visualization tools, are openly available on the Open Science Framework \citep{data_crowdsourced_2022}. This version of the dataset has been split into cross-validation folds based on participant.

\subsubsection{DreamerA and DreamerV}
\textbf{\textit{Dreamer}} is a multimodal dataset that includes electroencephalogram (EEG) and electrocardiogram (ECG) signals recorded during affect elicitation using audio-visual stimuli~\citep{dreamer}, captured with a 14-channel Emotiv EPOC headset. It consists of data recording from 23 participants, along with their self-assessments of affective states (valence, arousal, and dominance) after each stimulus~\citep{dreamer}. For our classification task, we focus on the arousal and valence labels, referred to as \textbf{\textit{DreamerA}} and \textbf{\textit{DreamerV}} respectively.

The dataset is publicly available \citep{data_dreamer_2017}, and we utilize the Torcheeg toolkit for preprocessing, including signal cropping and low-pass and high-pass filtering \citep{zhang_etal_2024}. Note that only EEG data is analyzed in this study, with ECG signals excluded. Labels for arousal and valence are binarized, assigning values below 3 to class 1 and values of 3 or higher to class 2, and has been split into cross-validation folds based on participant.

\subsubsection{STEW: Simultaneous Task EEG Workload}
\textbf{\textit{STEW}} comprises raw EEG recordings from 48 participants involved in a multitasking workload experiment~\citep{stew}. Additionally, the subjects' baseline brain activity at rest was recorded before the test. The data was captured using the Emotiv Epoc device with a sampling frequency of 128Hz and 14 channels, resulting in 2.5 minutes of EEG recording for each case. Participants were instructed to assess their perceived mental workload after each stage using a rating scale ranging from 1 to 9, and these ratings are available in a separate file. The dataset has been divided into cross-validation folds based on individual participants. Additionally, binary class labels have been assigned to the data, categorizing workload ratings above 4 as ``high'' and those below or equal to 4 as ``low''. We utilize these labels for our specific problem. STEW can be accessed upon request through the IEEE DataPort \citep{data_stew_2022}.

\subsection{Human Activity Recognition Datasets}

\subsubsection{Opportunity}

\textbf{\textit{Opportunity}} is a comprehensive, multi-sensor dataset designed for human activity recognition in a naturalistic environment~\citep{Opportunity}. Collected from four participants performing typical daily activities, the dataset spans six recording sessions per person: five unscripted ``Activities of Daily Living'' (ADL) runs, and one structured ``drill'' run with specific scripted activities. This dataset includes rich, multi-level annotations; however, for our analysis, we focus specifically on the locomotion classes, which consist of five primary categories: Stand, Walk, Sit, Lie, and Null (no specific activity detected).

Data collection includes 113 sensor channels from body-worn, object-attached, and ambient sensors. These channels capture essential information on body movements, object interactions, and environmental contexts through inertial measurement units, accelerometers, and switches. The variety and placement of these sensors allow for detailed examination of physical activities and transitions in a natural setting. To prepare the data for analysis, we segment it using a sliding window approach with a 100 time-step window and an overlap of 50 time steps. This segmentation enables the model to capture both the continuity of activities and subtle transitions, enhancing recognition accuracy across the locomotion classes. The dataset has been divided into cross-validation folds based on individual participants.

\subsubsection{PAMAP2: Physical Activity Monitoring Dataset}

\begin{figure}
\centering
     \includegraphics[trim=1cm 0cm 0cm 1cm, width=0.9\columnwidth]{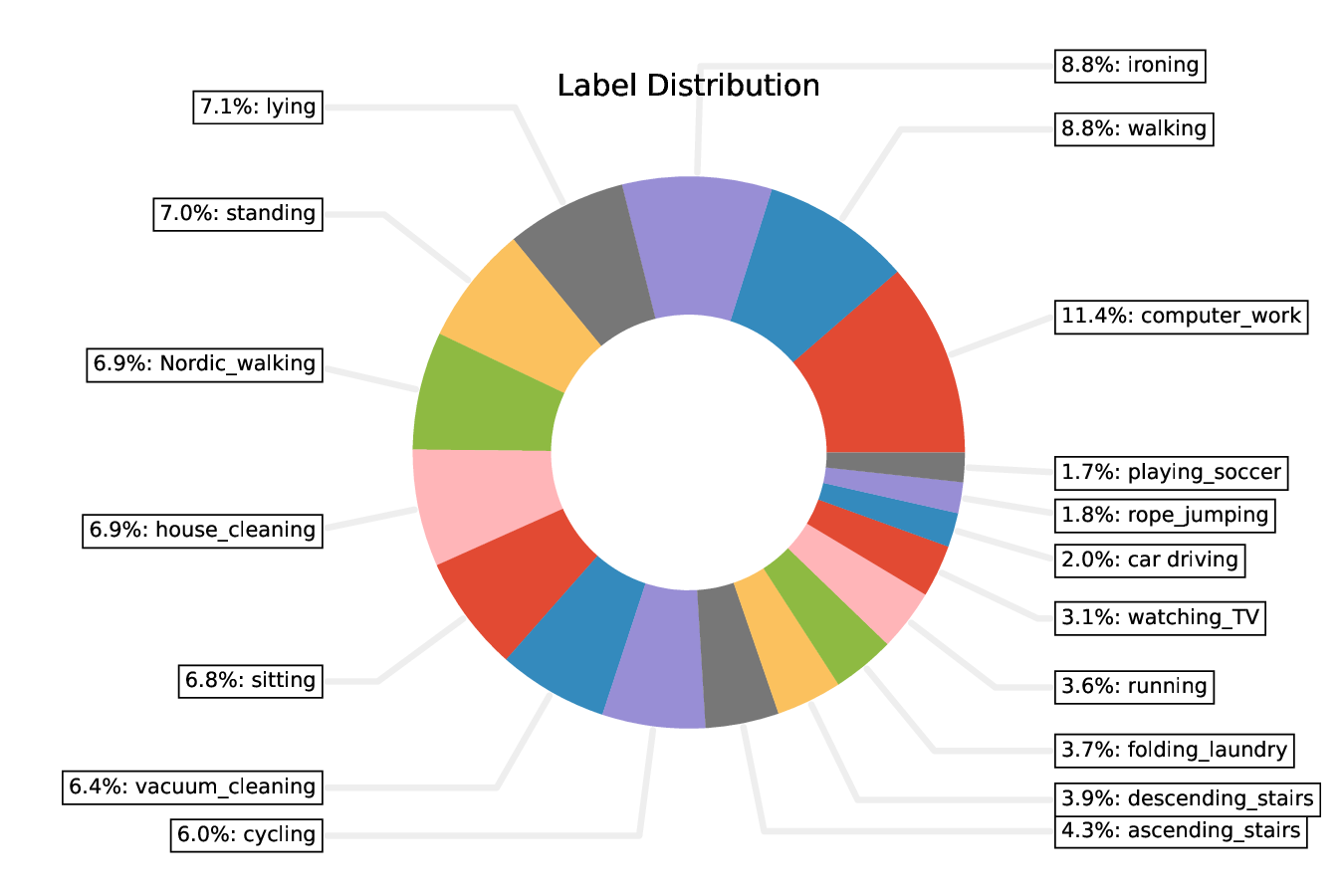}
     \caption{Distribution of activity categories for \textit{PAMAP2}.}
     \label{fig:distrib_PAMAP2}
\end{figure}

\textbf{\textit{PAMAP2}} is a collection of data obtained from three Inertial Measurement Units (IMUs) placed on the wrist of the dominant arm, chest, and ankle, as well as 1 ECG heart rate \citep{PAMAP2-2012}. The data was recorded at a frequency of 100Hz. The dataset includes annotated information about human activities performed by 9 subjects, each with their own unique physical characteristics. The majority of the subjects are male and have a dominant right hand. Notably, the dataset includes only one female subject (ID 102) and one left-handed subject (ID 108). In total, there are 18 different human activity classes represented in the dataset. Figure~\ref{fig:distrib_PAMAP2} provides a visual representation of the distribution of these activities across all the data. To ensure an unbiased evaluation, we divide the dataset into cross-validation folds based on the subjects.

\subsubsection{Skoda: Mini Checkpoint-Activity recognition dataset}

\textbf{\textit{Skoda}} captures 10 specific manipulative gestures performed in a car maintenance scenario. Its purpose is to investigate different aspects related to the gestures, such as fault resilience, performance scalability with the number of sensors, and power performance management. The dataset comprises 10 classes of manipulative gestures, which were recorded using 2x10 USB sensors positioned on the left and right upper and lower arm. The sensors have a high sample rate of approximately 98Hz, ensuring precise capturing of the movements.

\begin{figure}
\centering
     \includegraphics[trim=1cm 0cm 0cm 1cm, width=0.95\columnwidth]{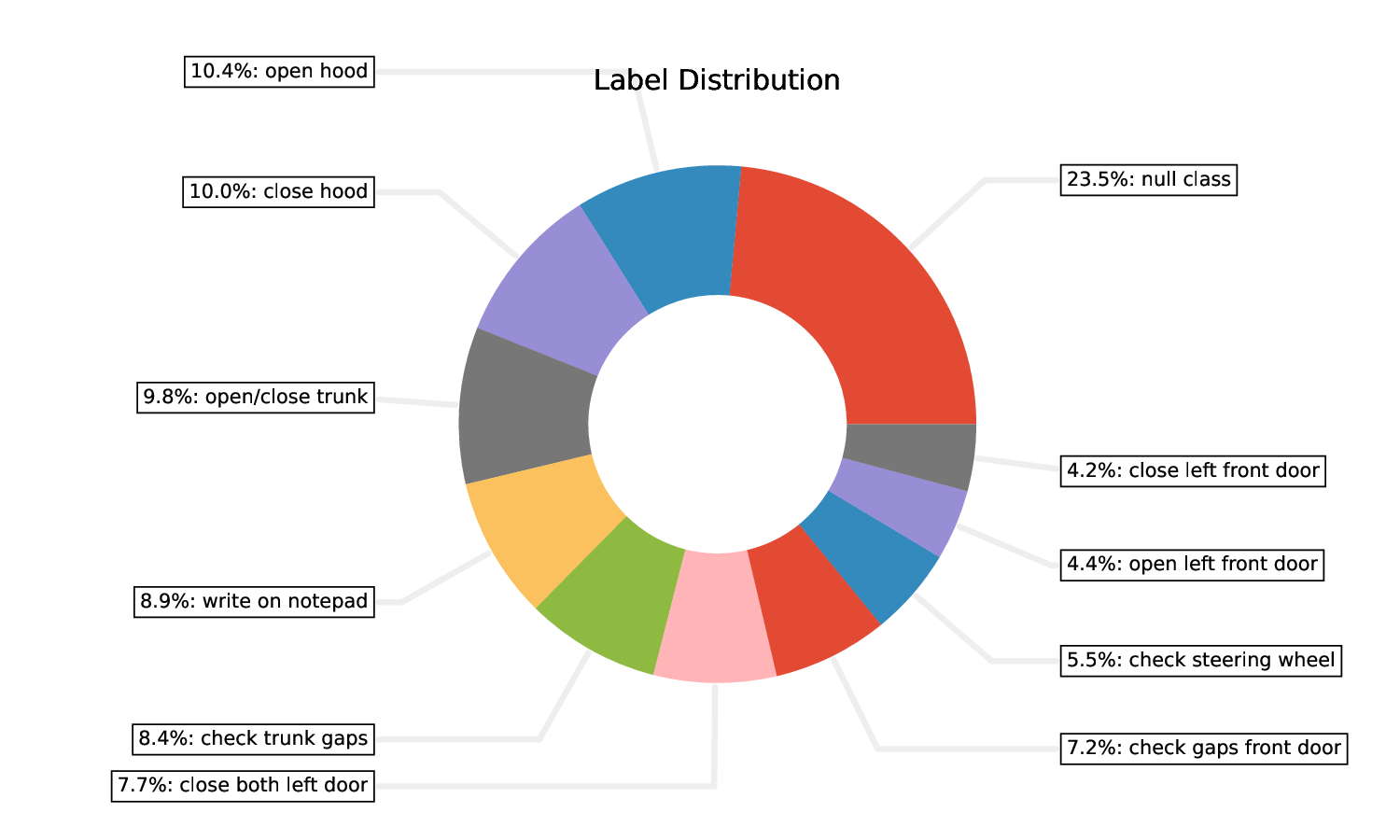}
     \caption{Distribution of activity categories for \textit{Skoda}.}
     \label{fig:distrib_SKODA}
\end{figure}

In terms of activities, the dataset includes 10 distinct manipulative gestures commonly performed during car maintenance (Figure~\ref{fig:distrib_SKODA}). The data was collected from a single subject, with each gesture being recorded 70 times. In total, the dataset offers around 3 hours of recording time, enabling thorough analysis of the gestures in various scenarios.

\subsubsection{UCIActivity}
\textbf{\textit{UCIActivity}} is a widely recognized benchmark for activity recognition research. It contains sensor readings from 30 participants performing six daily activities: walking, walking upstairs, walking downstairs, sitting, standing, and lying down. The data was collected using a Samsung Galaxy S2 smartphone mounted on the waist of each participant, with a sampling rate of 50 Hz \cite{UCI-HAR}. To keep the evaluation fair, we perform subject-wise cross-validation.

\subsubsection{USCActivity: USC human activity dataset}

\begin{figure}
\centering
     \includegraphics[trim=0cm 0cm 0cm 0cm, width=0.95\columnwidth]{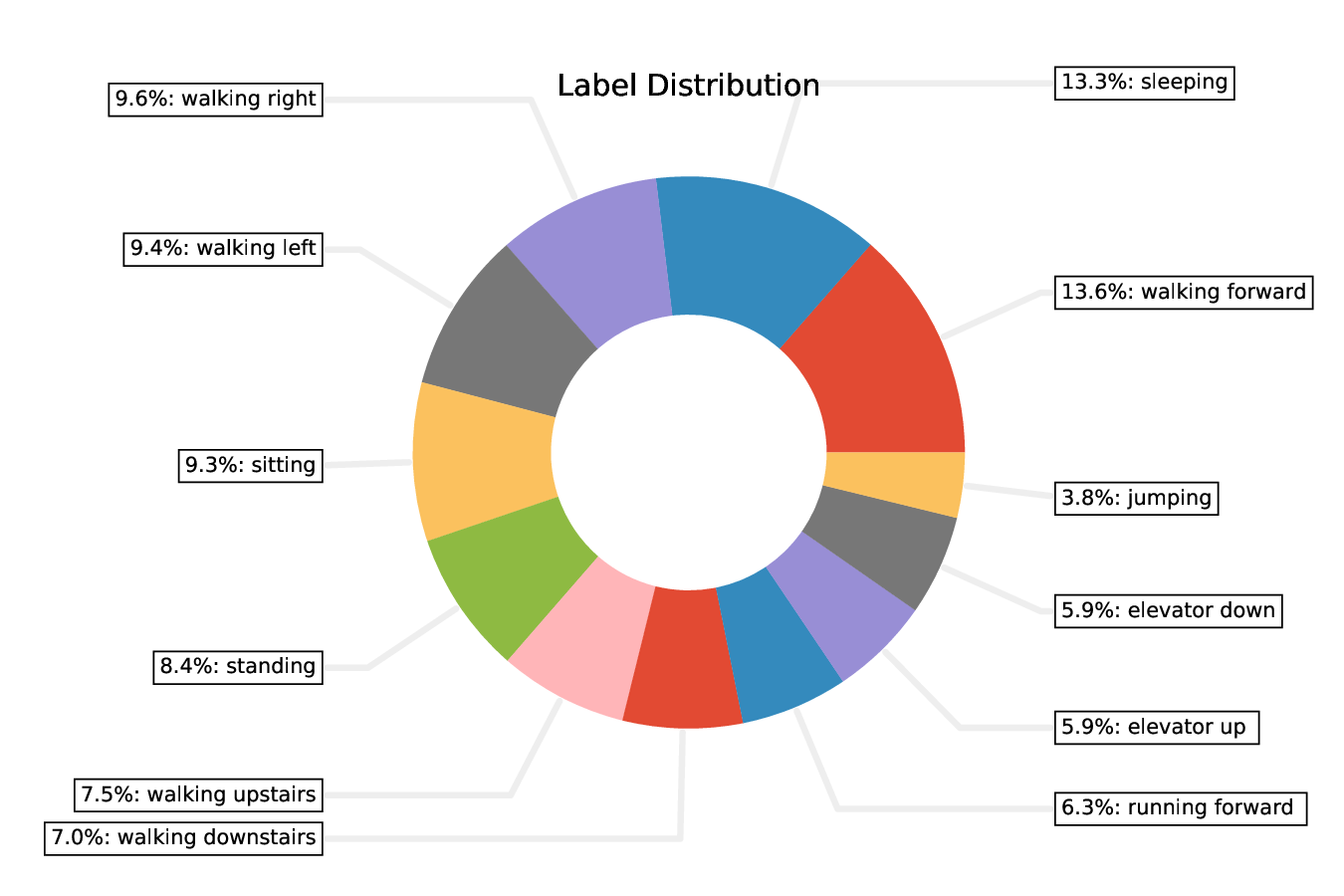}
     \caption{Distribution of activity categories for \textit{USC-HAD}.}
     \label{fig:distrib_USC}
\end{figure}

\textbf{\textit{USCActivity}} \citep{USC-HAD2012} consists of data collected from a Motion-Node device, which includes six readings from a body-worn 3-axis accelerometer and gyroscope sensor. The dataset contains samples from 14 male and female subjects with equal distribution (7 each) and specific physical characteristics and ages. The sensor data is sampled at a rate of 100 Hz, and each time-step in the dataset is labeled with one of 12 activity classes (Figure~\ref{fig:distrib_USC}).

The USCActivity dataset presents a challenge in learning feature representation and segmentation due to the placement of the sensors and the variability in activity classes. The data is collected from a single accelerometer and gyroscope reading obtained from a motion node attached to the subject's right hip. Therefore, this reading does not contribute significantly to the feature space transformation. Additionally, the activity classes involve various orientations, such as walking forward, left, or right, and even using the elevator up or down, which cannot be captured solely through accelerometer and gyroscope readings. Similar to other activity recognition datasets, we use subject-based cross-validation.

\subsubsection{WISDM and WISDM2: Wireless Sensor Data Mining}

\textbf{\textit{WISDM}} describes six daily activities collected in a controlled laboratory environment. 
The activities include \textit{Walking}, \textit{Jogging}, \textit{Stairs}, \textit{Sitting}, \textit{Standing}, and \textit{Lying Down}, recorded from 36 users using a cell phone placed in their pocket. 
The data is sampled at a rate of 20 Hz, resulting in a total of 1,098,207 samples across 3 dimensions \citep{WISDM}.

\textbf{\textit{WISDM2}} extends the original WISDM dataset by collecting data in real-world environments using the Actitracker system. 
This system was designed for public use and provides a more extensive collection of sensor readings from users performing the same six activities. 
The dataset contains 2,980,765 samples with three dimensions, and the data was recorded from a larger and more diverse set of participants in naturalistic settings, offering a valuable resource for real-world activity recognition \citep{WISDM2}. Both WISDM and WISDM2 are split based on subjects.

\subsection{Counts}

\subsubsection{Pedestrian}

\textbf{\textit{Pedestrian}} represents hourly pedestrian counts at 82 locations in Melbourne, Australia between 2009 and 2022 \citep{data_pedestrian_2022}. The processed dataset consists of $189{,}621$ (univariate) time series, each of length 24. The task is to identify location based on the time series of counts. The dataset has been split into stratified random cross-validation folds.

\subsubsection{Traffic}

\textbf{\textit{Traffic}} consists of hourly traffic counts at various locations in the state of NSW, Australia \citep{data_traffic_2023}. The processed dataset contains $1{,}460{,}968$ (univariate) time series, each of length $24$. The task is to predict the day of the week based on the time series of counts. The dataset has been split into stratified random cross-validation folds.

\subsection{Other}

\subsubsection{FordChallenge}

\textbf{\textit{FordChallenge}} is obtained from Kaggle and consists of data from 600 real-time driving sessions, each lasting approximately 2 minutes and sampled at 100ms intervals~\citep{stayalert}. These sessions include trials from 100 drivers of varying ages and genders. The dataset contains 8 physiological, 11 environmental, and 11 vehicular measurements, with specific details such as names and units undisclosed by the challenge organizers. Each data point is labeled with a binary outcome: 0 for ``distracted'' and 1 for ``alert.'' The objective of the challenge is to design a classifier capable of accurately predicting driver alertness using the provided physiological, environmental, and vehicular data.

\subsubsection{LenDB}

\textbf{\textit{LenDB}} consists of seismograms recorded from multiple different seismic detection networks from across the globe \citep{data_lendb_2020,magrini_etal_2020}. The processed dataset consists of $1{,}244{,}942$ multivariate time series, with $3$ channels, each of length $540$, with two classes: earthquake and noise. This version of the dataset has been split into cross-validation folds based on seismic detection network (i.e., such that seismograms for a given network do not appear in both a training and validation fold).


\section{Baseline Results} \label{sec-results}

\subsection{Models}

We provide baseline results on the {\monster} datasets for a number of key models. In particular, we provide results for four deep learning models: ConvTran \citep{ConvTran}, FCN \citep{Wang2017}, HInceptionTime \citep{Ismail-fawaz2022}, and TempCNN \citep{Pelletier2019}. We include results for two more `traditional', specialised methods for time series classification: {\hydra} \citep{dempster_etal_2023}, and {\quant} \citep{dempster_etal_2024b}. We also include results for a standard, `off the shelf' classifier---extremely randomised trees \citep{geurts_etal_2006}---to act as a na\"{i}ve baseline.

\textbf{{\fcn}} is a fully convolutional neural network. It consists of three temporal convolutional layers (one-dimensional convolutional layers that convolve along the time series), followed by a global average pooling layer and finally the softmax classification layer~\citep{Wang2017}. The convolutional layers have 128, 256, and 128 filters of length 8, 5, and 3, respectively.

\textbf{{\tempcnn}} is a light-weight temporal convolutional neural network originally designed for land cover classification from time series of satellite imagery~\citep{Pelletier2019}. It consists of three temporal convolutional layers followed by a fully connected layer. Each convolutional layer has 64 filters of length 5 and the fully-connected layer has 256 units. 

\textbf{{\hinception}} (Hybrid-InceptionTime) is an ensemble of five Hybrid-Inception (H-Inception) models, each with a different random weight initialisation~\citep{Ismail-fawaz2022}. An H-Inception model consists of a set of 17 hand-crafted filters combined with six Inception modules. The hand-crafted filters are sets of convolutional filters designed to detect peaks, and both increasing and decreasing trends. The hand-crafted filters range in length from 2 to 96 and are applied in parallel with the first inception module to the input time series. Inception modules combine convolutions with filter lengths of 10, 20 and 40, max pooling and bottleneck layers. Each set of convolutions and the max pooling layer have 32 filters thus each inception module has 128 filters. The resulting network has a small number of parameters and a large receptive field~\citep{Fawaz2019a}.

\textbf{ConvTran} is a deep learning model for multivariate time series classification (TSC) that combines convolutional layers with transformers to effectively capture both local patterns and long-range dependencies~\citep{ConvTran}. It addresses the limitations of existing position encoding methods by introducing two novel techniques: tAPE (temporal Absolute Position Encoding) for absolute positions and eRPE (efficient Relative Position Encoding) for relative positions. These encodings, integrated with disjoint temporal and channel-wise convolutions~\citep{Disjoin-CNN}, allow ConvTran to capture both temporal dependencies and correlations between the channels.

\textbf{{\hydra}} involves transforming input time series using a set of random convolutional kernels arranged into groups, and `counting' the kernel representing the closest match with the input time series in each group. The counts are then used to train a ridge regression classifier \citep{dempster_etal_2023}. Here, we use the variant of {\hydra} presented in \citet{dempster_etal_2024}, which integrates the {\hydra} transform into the process of fitting the ridge regression model, and all computation is performed on GPU.

\textbf{{\quant}} involves recursively dividing the input time series in half, and computing the quantiles for each of the resulting intervals (subseries) \citep{dempster_etal_2024b}. The computed quantiles are used to train an extremely randomised trees classifier. {\quant} acts on the original input time series, the first and second derivatives, and the Fourier transform. Here, we use the variant of {\quant} presented in \citet{dempster_etal_2024}, which uses pasting to `spread' the extremely randomised trees over the dataset.

\textbf{Extremely Randomised Trees} (`ET') is a well-established classifier, using an ensemble of decision trees where a random subset of features and split points is considered at each node, with the feature/split chosen which minimises log loss \citep{geurts_etal_2006}. Here, we use the same setup as for {\quant}, but remove the {\quant} transform, so that ET is training directly on the `raw' time series data (rather than the {\quant} features). ET serves as a `na\"{i}ve' baseline reference point for the other models.

The four deep learning models are trained using the Adam optimiser~\citep{Kingma2015} and a batch size of 256 for a maximum of 100 epochs. The one exception is HInceptionTime with the AudioMNIST dataset, which used a batch size of 64 to enable it to fit in the GPU memory. For all datasets, we implement early stopping and select the best epoch found as the final model, using a validation set obtained by randomly selecting 10\% of the training dataset. Training time on each fold is limited to approximately 24 hours or one epoch, whichever is longer.

We provide results for {\zo}, log loss, and training time. (We also provide learning curves for {\hydra} and {\quant}: see Appendix \ref{sec-appendix-curves-hydra-quant}). Each method is evaluated on each dataset using 5-fold cross-validation, using predefined cross-validation folds. (Note that both Quant and ET are unable to train on one of the folds of the WISDM dataset, due a limitation of the ET implementation where there is a single example of a given class.)  These results serve as an initial survey on the relative performance of different methods on the {\monster} datasets, to serve as a reference point for future work on large time series classification tasks.

\subsection{Summary}

\begin{figure}%
    \centering%
    \includegraphics[width=0.85\linewidth]{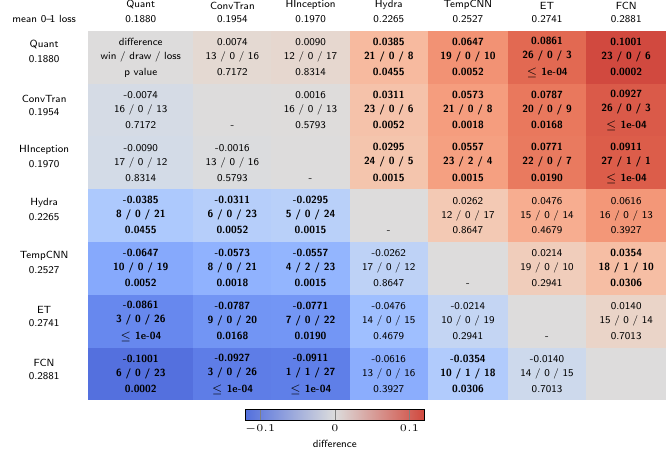}%
    \caption{Multi-comparison matrix showing mean {\zo} and pairwise differences.}%
    \label{fig-mcm}%
\end{figure}

The multi-comparison matrix (MCM) in Figure~\ref{fig-mcm} shows mean {\zo} as well as pairwise differences and win/draw/loss for the baseline methods over all 29 {\monster} datasets \citep[see][]{ismailfawaz_etal_2023b}.

Figure \ref{fig-mcm} shows that {\quant} achieves the lowest overall mean {\zo}, slightly lower than that of ConvTran, although both ConvTran and HInceptionTime have lower {\zo} on more datasets (16 vs 13 and 17 vs 12 respectively). Hydra has higher overall mean {\zo} than HInceptionTime, but lower than TempCNN, ET, or FCN. TempCNN, ET, and FCN all have higher average {\zo}, due in large part to poor performance on the audio datasets: see Section \ref{subsection-by-category}.

\subsection{By Category} \label{subsection-by-category}

\begin{figure}%
    \centering%
    \includegraphics[width=\linewidth]{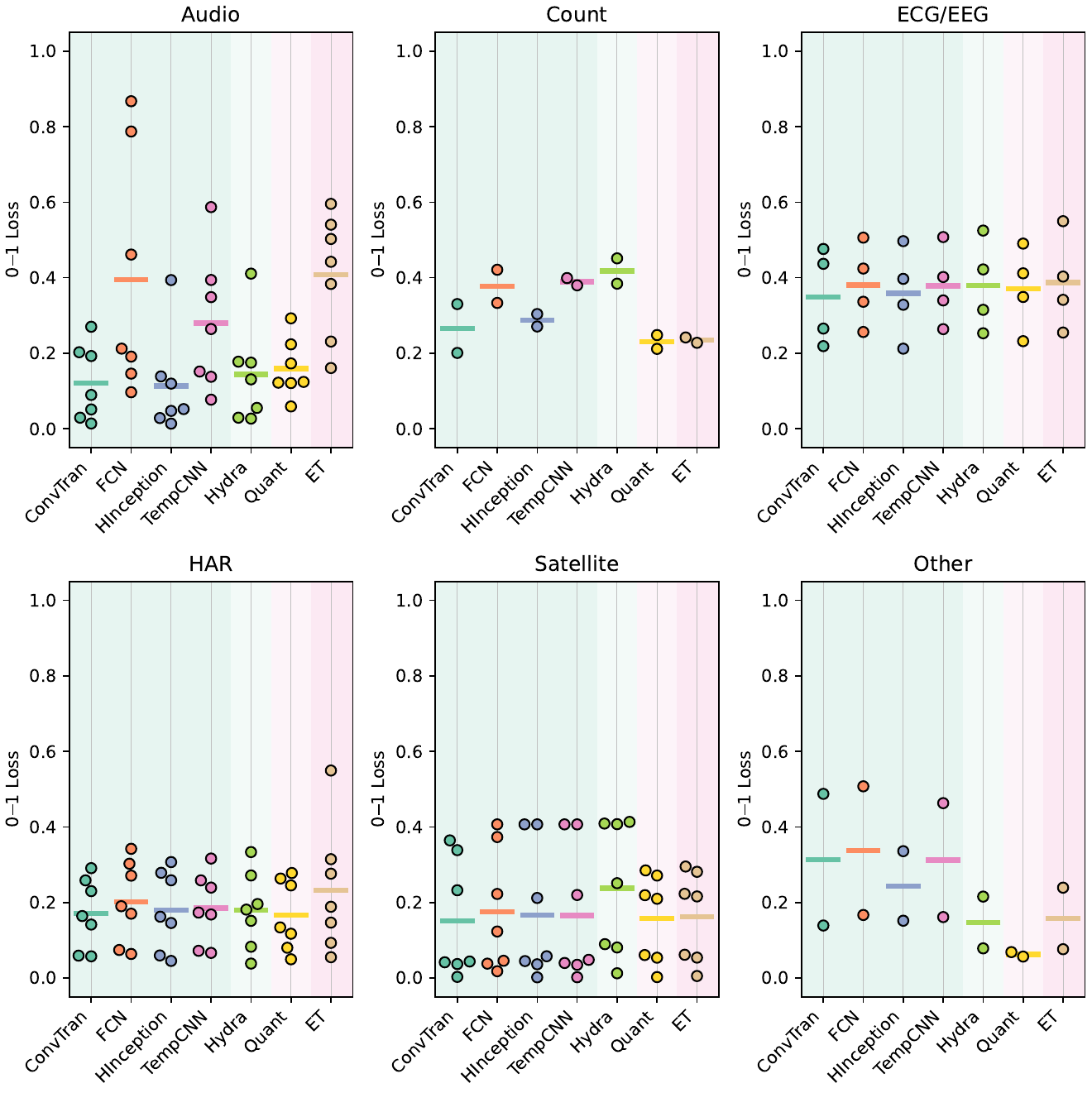}%
    \caption{{\zo} by category.}%
    \label{fig-by-category}%
\end{figure}

Figure \ref{fig-by-category} shows the {\zo} for each method on each dataset, organised by category (Audio, Count, ECG/EEG, HAR, Satellite, and Other). Each point represents a single dataset. The horizontal bars represent mean {\zo} for each classifier within each category. Figure~\ref{fig-by-category} shows that while for some categories the {\zo} for different methods is broadly similar, for other categories there are considerable differences.

In particular, ConvTran, HInceptionTime, Hydra and {\quant} all achieve relatively low {\zo} on the audio datasets, while ET, TempCNN, and especially FCN have much higher {\zo}. ET, {\quant}, and (to a lesser extent) ConvTran and HInception achieve relatively low {\zo} on the count datasets. {\quant} and (to a lesser extent) ET and {\hydra} achieve relatively low {\zo} on the `other' datasets.

In contrast, mean {\zo} for ECG/EEG, HAR, and Satellite is broadly similar, with significant spread within the results for each method.

Interestingly, it is only on the audio datasets, and to some extent the HAR datasets, that our na\"{i}ve baseline, ET, appears to be meaningfully `worse' than the deep learning or specialised time series classification methods. ET achieves similar results to {\quant} on a number of datasets, which is not surprising, as the `raw' time series are similar to a subset of the features used in {\quant}.

We speculate that the poor {\zo} for FCN and TempCNN on the audio datasets in particular may be related to the small receptive field of these models (relative to the relatively long time series in the audio datasets). With a small receptive field, these models are in effect limited to high-frequency features in the data.

The satellite datasets show some interesting extremes. All methods except for {\quant} and ET performed poorly on the \textit{S2Agri} $10\%$ datasets. In contrast, all methods achieved very low {\zo} on \textit{LakeIce} as this dataset has strong temporal and spatial correlations between samples that could not be accounted for when splitting the data into folds.

\subsection{Computational Efficiency}

\subsubsection{Training Time}

Table \ref{table-training-time} shows total training time for each of the baseline methods, separated into methods using GPU and methods using CPU. This represents the total training time over all 29  {\monster} datasets (where the time for each dataset is the average training time across the five cross-validation folds). 

\newcolumntype{C}[1]{>{\centering\arraybackslash}m{#1}}%
\begin{table}
    \centering
    \scriptsize
    \caption{Total Training Time}
    \label{table-training-time}
    \begin{tabular}{C{15mm}C{15mm}C{15mm}C{15mm}C{15mm}C{15mm}C{15mm}}
    \toprule
    \multicolumn{5}{c}{\textbf{GPU}} & \multicolumn{2}{c}{\textbf{CPU}} \\
    \cmidrule(lr){1-5}\cmidrule(lr){6-7}
    Hydra & ConvTran & TempCNN & FCN & HInception & ET & Quant \\
    \cmidrule(lr){1-5}\cmidrule(lr){6-7}
    47m 44s & 5d 6h & 2d 9h & 2d 12h & 6d 6h & 5h 10m & 20h 10m \\
    \bottomrule
    \end{tabular}
\end{table}

The five methods trained using GPUs were each trained using a single GPU, either an Ampere A100 SMX4 with 80GB RAM, or an Ampere A40 with 48GB RAM. Table~\ref{table-training-time} shows that among these methods, {\hydra} is by far the fastest, taking less than 50 minutes to train over all 29 {\monster} datasets, almost $70\times$ faster than the next-fastest GPU method (TempCNN). HInceptionTime is the least efficient method, requiring approximately a week of training time, corresponding to approximately one month total training time across all five cross-validation folds. (We note that there is a variant of {\hinception}, LITETime, with significantly fewer parameters which requires less than half of the training time of {\hinception}: \citet{ismailfawaz_etal_2023}.)

Although not directly comparable to methods using GPU, {\quant} requires approximately 20 hours of training time (using 4 CPU cores). ET requires only approximately a quarter of this (5 hours), due to the smaller number of features used to train the classifier.

\subsubsection{Parameter Counts}

\begin{table}
    \centering
    \scriptsize
    \caption{Number of Parameters}
    \label{table-parameters}
    \begin{tabular}{C{5mm}C{18mm}C{18mm}C{18mm}C{18mm}C{18mm}C{18mm}}
    \toprule
    {} & ConvTran & FCN & HInception & TempCNN & Hydra\textsuperscript{\textdagger} & Quant\textsuperscript{\textdaggerdbl} \\
    \midrule
    \textit{min} & \shortstack{27,039 \\ Traffic} & \shortstack{264,962 \\ CornellWhale} & \shortstack{869,570 \\ CornellWhale} & \shortstack{424,649 \\ Tiselac} & \shortstack{6,144 \\ FordChallenge} & \shortstack{275 \\ CrowdSourced} \\
    \midrule
    \textit{max} & \shortstack{486,941 \\ Opportunity} & \shortstack{380,037 \\ Opportunity} & \shortstack{1,420,145 \\ Opportunity} & \shortstack{786,444,426 \\ AudioMNIST} & \shortstack{167,936 \\ Pedestrian} & \shortstack{379,112 \\ Traffic} \\
    \bottomrule
    \end{tabular}
    \newline
    {\textdagger} num. parameters in ridge classifier; {\textdaggerdbl} median num. leaves per tree
\end{table}

Table \ref{table-parameters} shows total number of parameters for each of the baseline methods. For each method the table shows the minimum and maximum number of parameters and the corresponding dataset. The number of parameters for both FCN and HInceptionTime is reasonably stable, with the largest model 1.4 and 1.6 times that of the smallest model, respectively. However, the number of parameters in the TempCNN models vary greatly, with the largest model being over $1{,}800$ times the size of the smallest one. While the total number of parameters for all the deep learning methods is dependent on the number of classes and channels, the FCN and HInceptionTime architectures both include a global average pooling layer, so the parameter count is independent of the length of the time series. However, TempCNN does not use global pooling and so its parameter count is highly dependent on the length of the time series. For {\hydra}, we have used the number of parameters for the ridge classifier, and for {\quant} we have used the median number of leaf nodes, although these are not directly comparable to the number of trainable parameters in the deep learning models.

\subsection{Pairwise Comparisons}

\begin{figure}%
    \centering%
    \includegraphics[width=\linewidth]{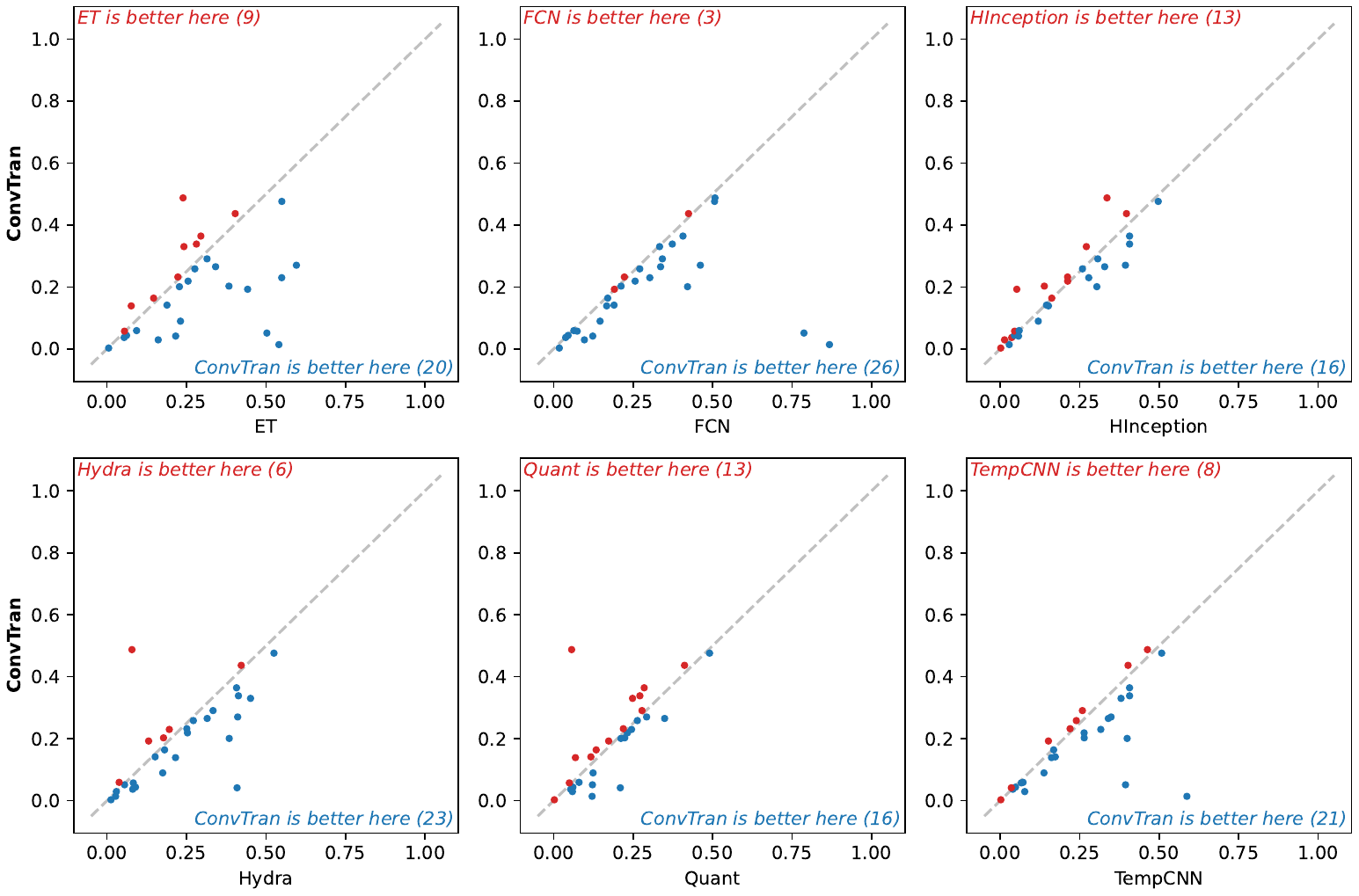}%
    \caption{Pairwise \textbf{{\zo}} for ConvTran.}%
    \label{fig-pairwise-zo}%
\end{figure}%

\begin{figure}%
    \centering%
    \includegraphics[width=\linewidth]{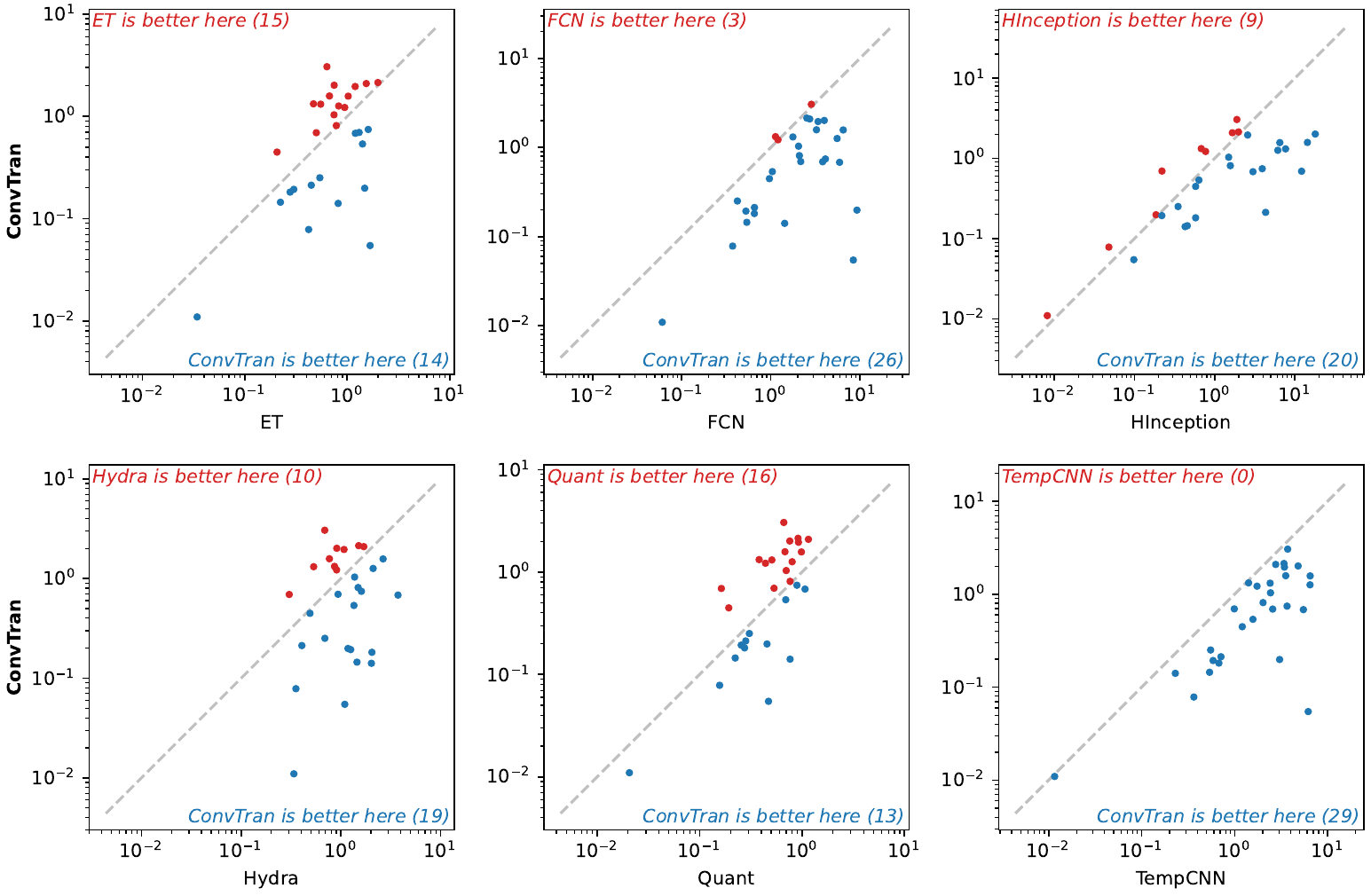}%
    \caption{Pairwise \textbf{log-loss} for ConvTran.}%
    \label{fig-pairwise-log}%
\end{figure}

\begin{figure}%
    \centering%
    \includegraphics[width=\linewidth]{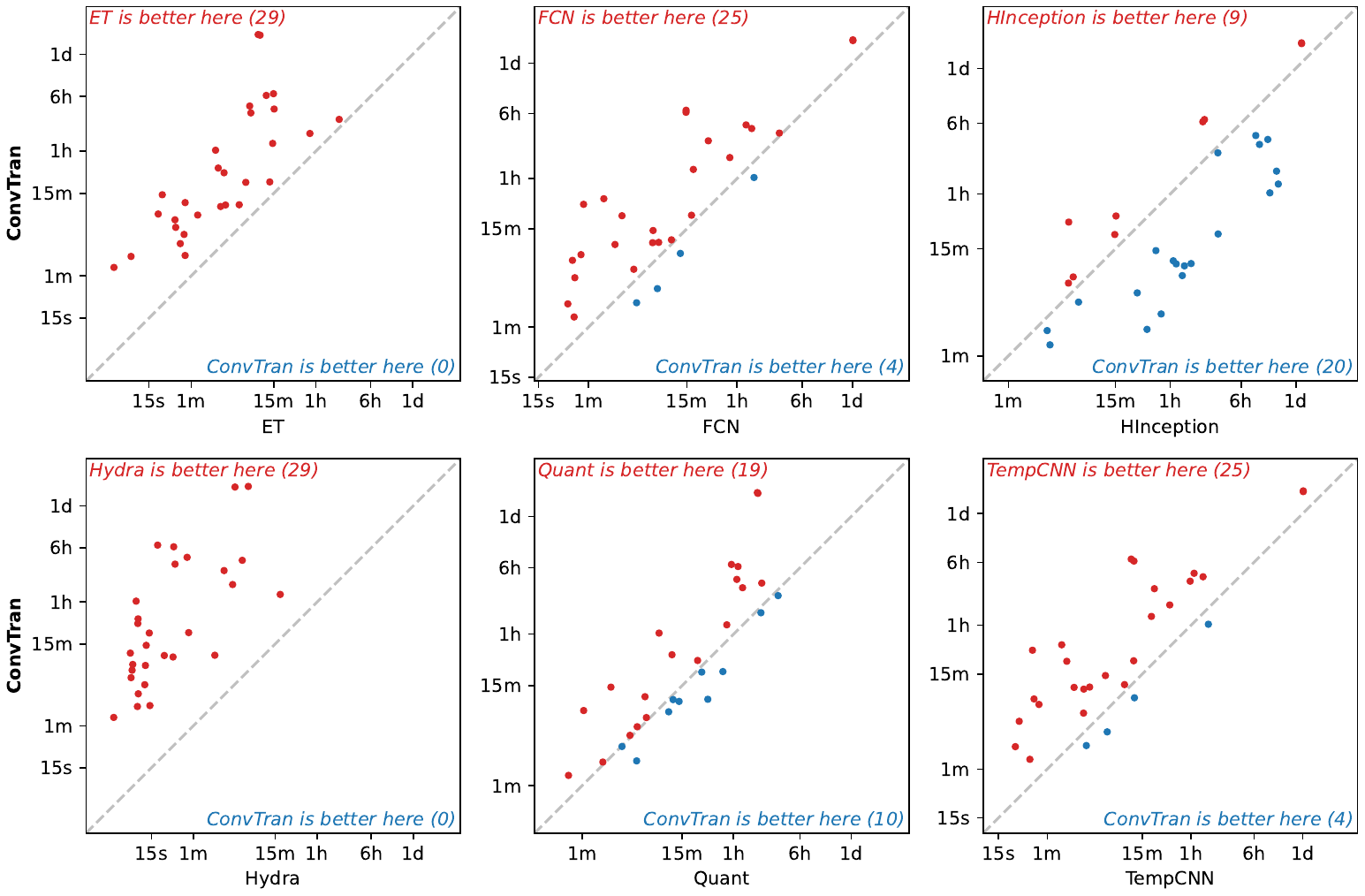}%
    \caption{Pairwise \textbf{training time} for ConvTran.}%
    \label{fig-pairwise-training-time}%
\end{figure}

Figures \ref{fig-pairwise-zo}, \ref{fig-pairwise-log}, and \ref{fig-pairwise-training-time} show the pairwise {\zo}, log loss, and training time for ConvTran versus each of the other baseline methods. (Full pairwise results for all methods and metrics are provided in the Appendix.)  Figure \ref{fig-pairwise-zo} shows that ConvTran achieves broadly similar {\zo} on most datasets compared to {\quant}, HInceptionTime, and {\hydra}, although ConvTran achieves noticeably lower {\zo} than {\hydra} on one dataset (and both {\hydra} and {\quant} achieve significantly lower {\zo} than ConvTran on one dataset).

While ConvTran achieves similar {\zo} to FCN and TempCNN on most datasets, ConvTran achieves considerably lower {\zo} on a small number of datasets. As noted above, these include the audio datasets, where FCN and TempCNN appear to struggle relative to the other methods.

Figure \ref{fig-pairwise-log} shows a slightly different picture in terms of log loss. ConvTran is fairly evenly matched to {\quant} (and ET) in terms of the number of datasets on which each method achieves lower log loss, although there is a considerable spread in values (i.e., they are not closely correlated). ConvTran achieves lower log loss on more datasets compared to {\hydra}, although {\hydra} does achieve lower log loss on 10 datasets, which is somewhat surprising, as {\hydra} takes no account of log loss in training. ConvTran achieves lower log loss than FCN, HInceptionTime, and TempCNN on most datasets.

Figure \ref{fig-pairwise-training-time} shows that ConvTran is significantly faster than HInceptionTime, but slower than FCN or TempCNN, on most datasets. ConvTran is marginally faster than {\quant} on a number of datasets, although the timings are not directly comparable (given that ConvTran uses GPU whereas {\quant} is limited to CPU). Both ET and {\hydra} are faster than ConvTran on all datasets (reflecting the overall differences in training time shown in Table \ref{table-training-time}).


\section{Conclusion} \label{sec-conclusion}

We present {\monster}, a new benchmark collection of large datasets for time series classification. The field of time series classification has become focused on smaller datasets. This has resulted in state-of-the-art methods being optimised for low average {\zo} over a large number of small datasets, has insulated the field from engaging with the challenges of learning from large quantities of data, and has artificially disadvantaged low-bias methods such as deep neural network models in benchmarking comparisons.

We hope that {\monster} encourages the field to engage with the challenges related to learning from large quantities of time series data. We hope that {\monster} will help better reflect the broader task of time series classification and improve relevance for real-world time series classification problems. We believe there is enormous potential for new research based on much larger datasets.

\impact{We present a new benchmark of 29 large datasets for time series classification. This could potentially have a large impact on the field, as these datasets are significantly larger than those currently used for benchmarking and evaluation. This should allow for training lower-bias, more complex models, with greater relevance and more direct applicability to large-scale, real-world time series classification problems. On the other hand, learning from larger quantities of data requires proportionally more computational time and resources. As such, it is important to always keep in mind the balance between computational expense and real-world relevance. There are also potential risks associated with the misuse of improved methods for time series classification in monitoring and surveillance, although we do not feel that there is any significant direct risk associated with this work.}

\acks{This work was supported by an Australian Government Research Training Program Scholarship, and the Australian Research Council under award DP240100048. The authors would like to thank, in particular, Professor Eamonn Keogh, Professor Tony Bagnall, and all the people who have contributed to the UCR and UEA time series classification archives. The authors also thank Raphael Fischer for trialling our methods and datasets and providing invaluable feedback.}

\vskip 0.2in
\bibliography{bibliography}

\clearpage
\appendix

\section{Additional Results}


\subsection{{\zo}}

\begin{figure}[h]%
    \centering%
    \includegraphics[width=0.85\linewidth]{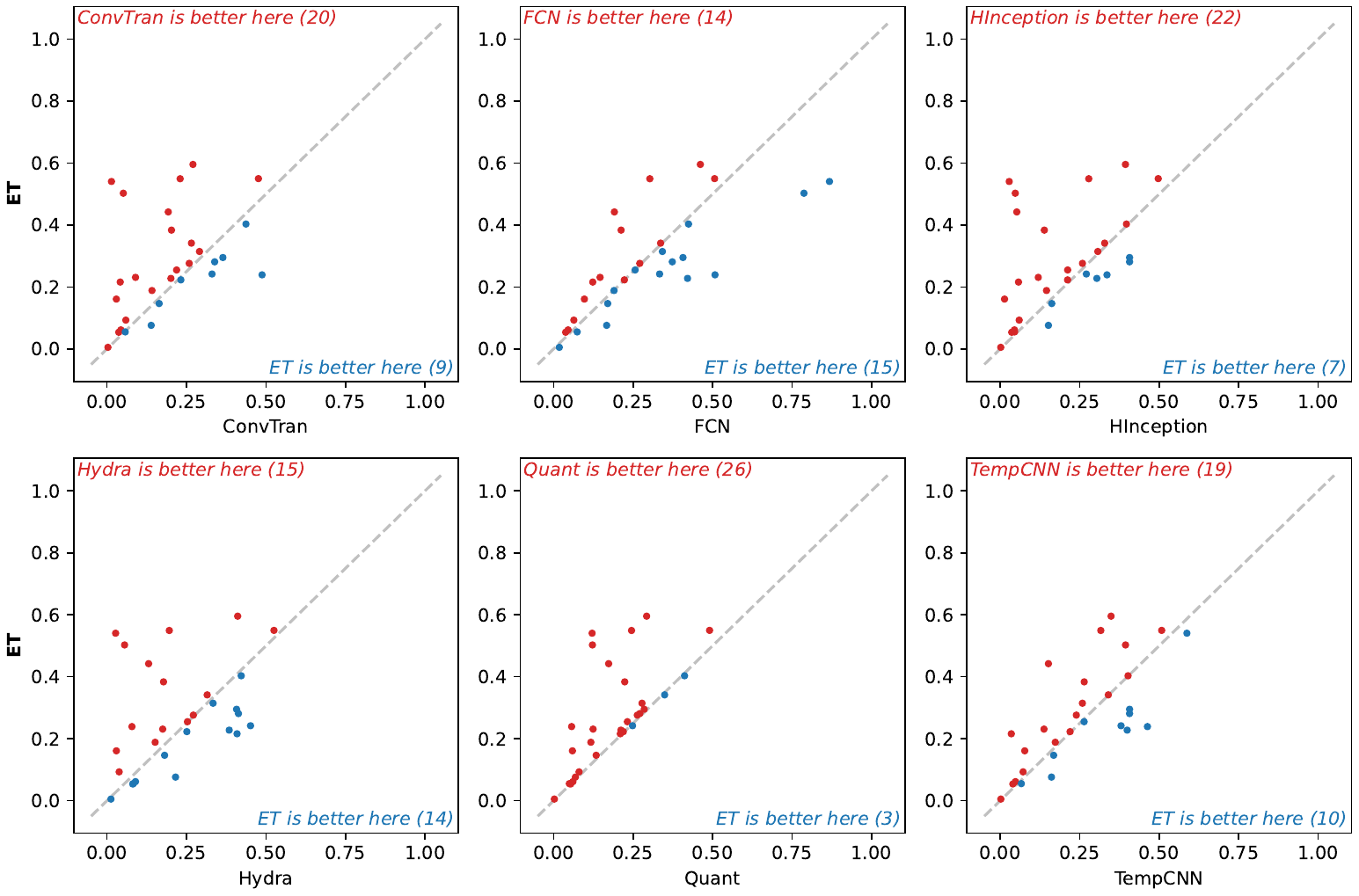}%
    \caption{Pairwise results ({\zo}) for ET.}%
\end{figure}%

\begin{figure}[h]%
    \centering%
    \includegraphics[width=0.85\linewidth]{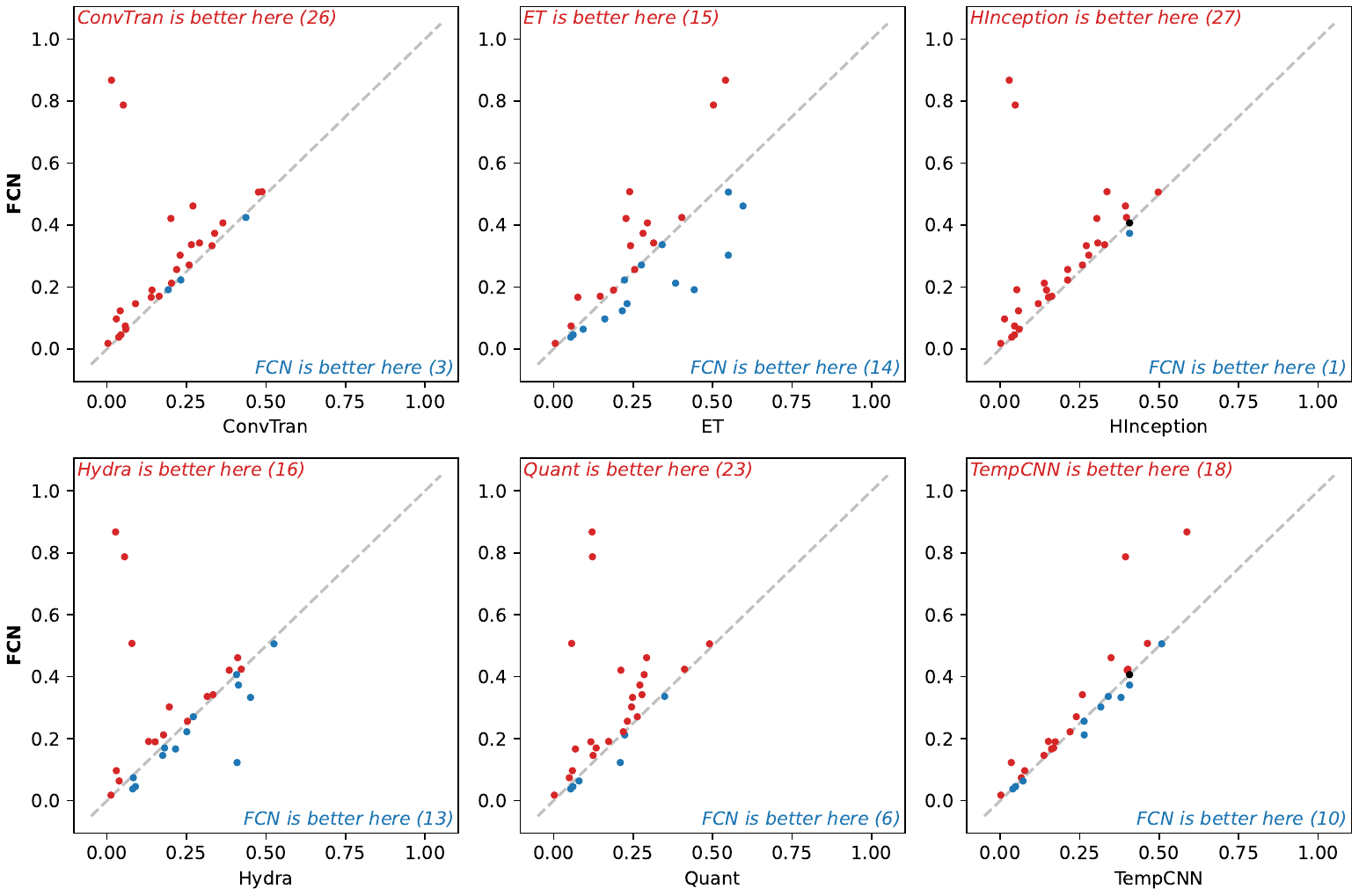}%
    \caption{Pairwise results ({\zo}) for FCN.}%
\end{figure}%

\begin{figure}[h]%
    \centering%
    \includegraphics[width=0.85\linewidth]{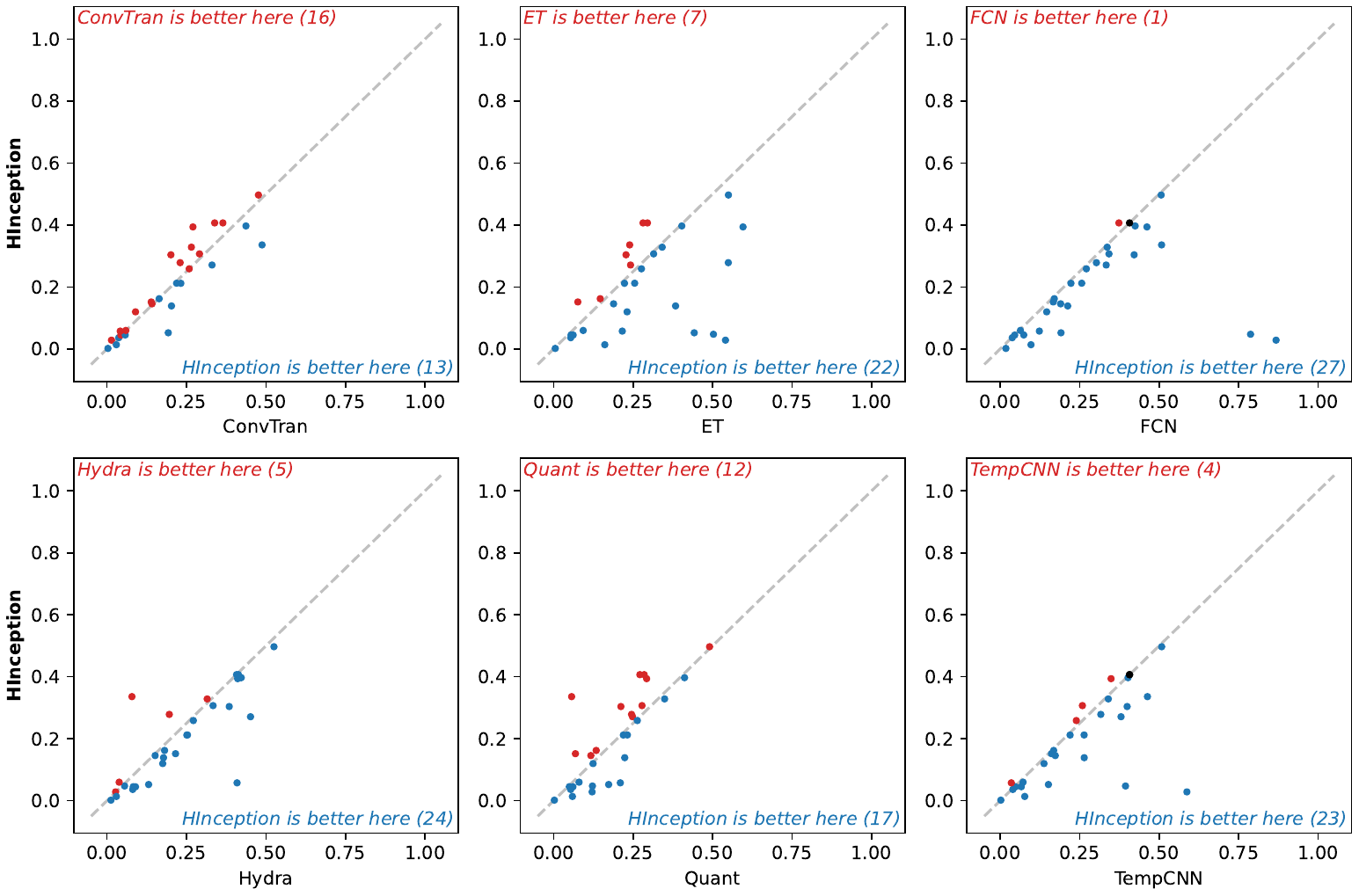}%
    \caption{Pairwise results ({\zo}) for HInceptionTime.}%
\end{figure}%

\begin{figure}[h]%
    \centering%
    \includegraphics[width=0.85\linewidth]{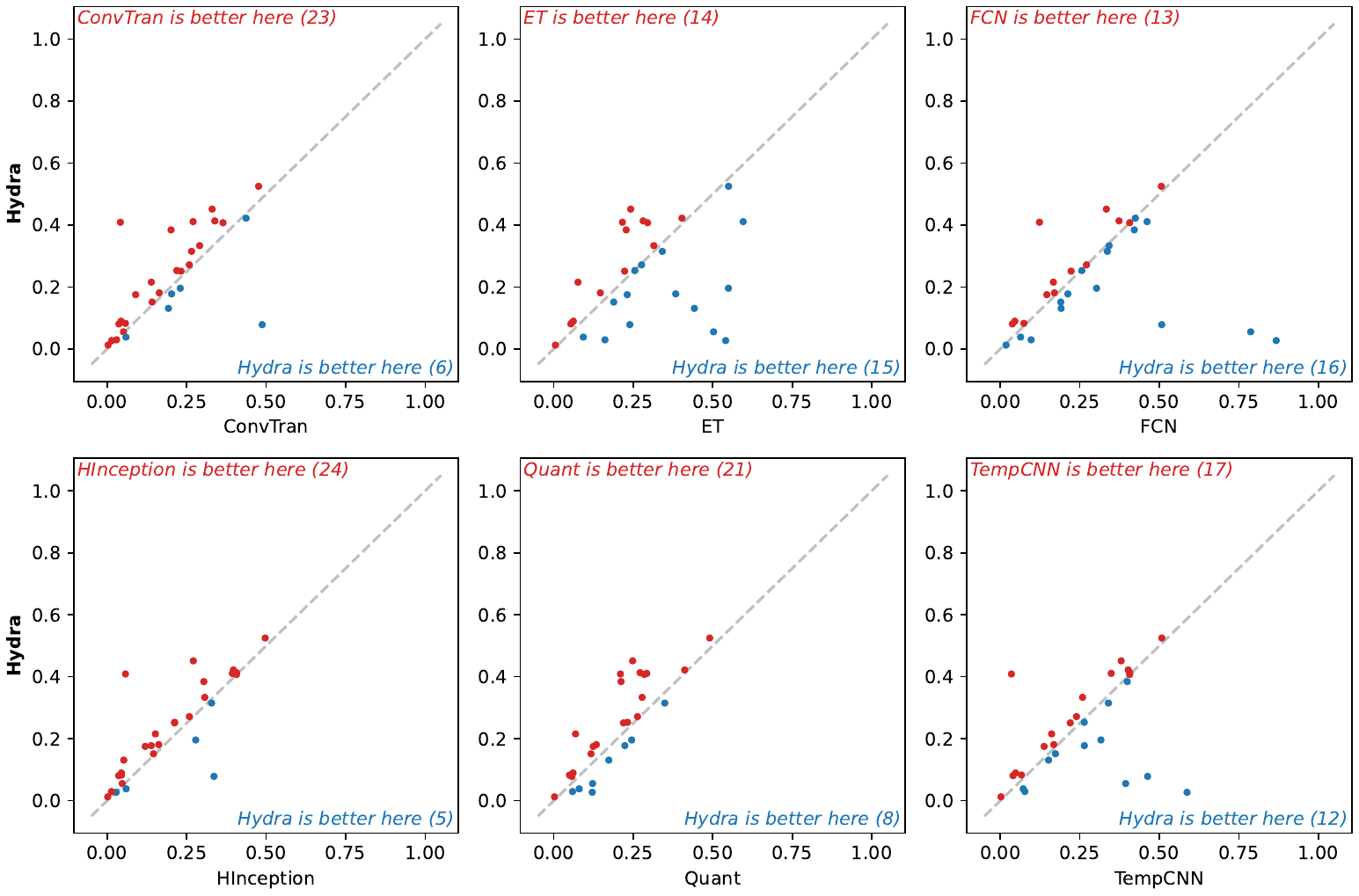}%
    \caption{Pairwise results ({\zo}) for {\hydra}.}%
\end{figure}%

\begin{figure}[h]%
    \centering%
    \includegraphics[width=0.85\linewidth]{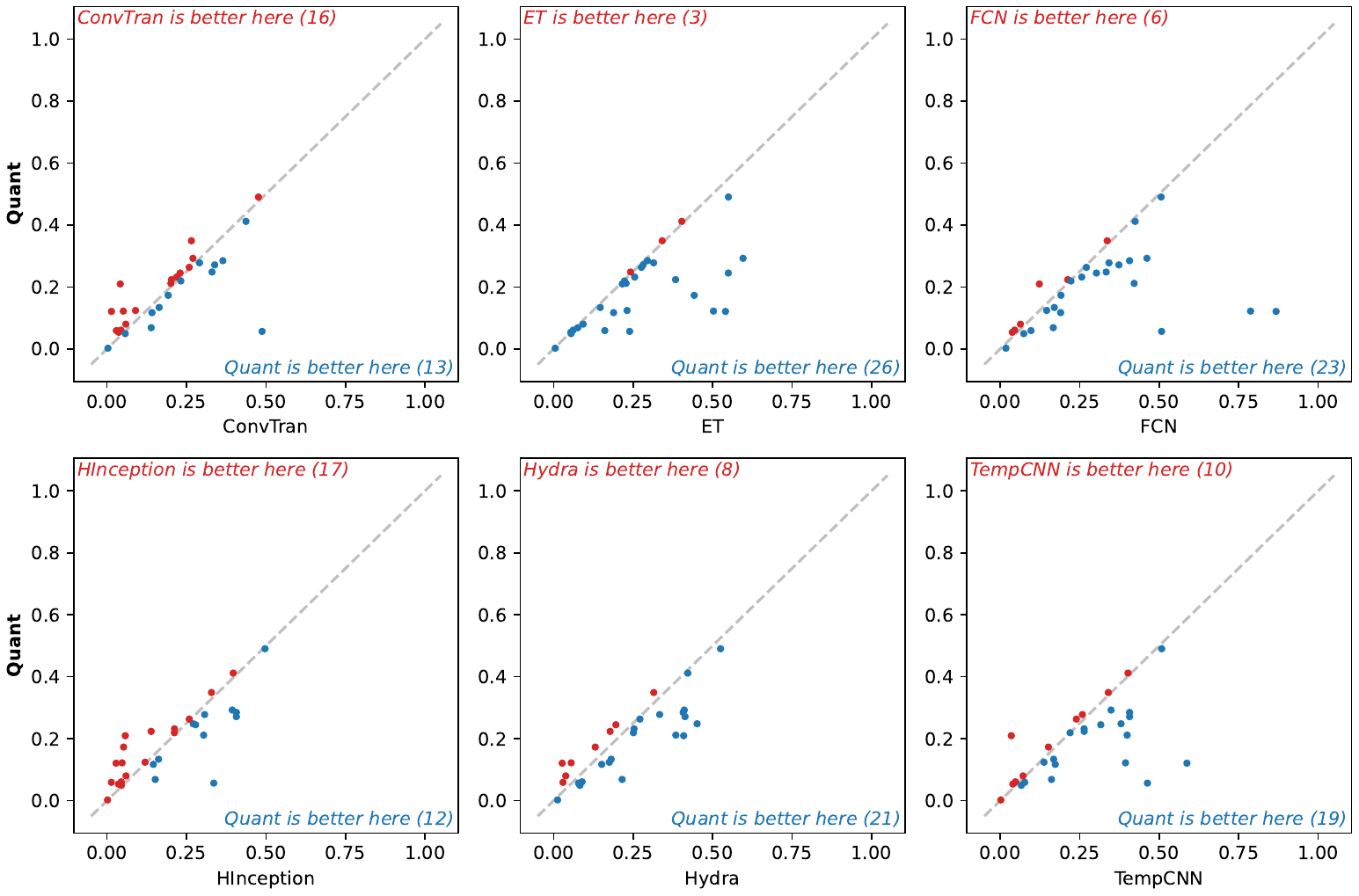}%
    \caption{Pairwise results ({\zo}) for {\quant}.}%
\end{figure}%

\begin{figure}[h]%
    \centering%
    \includegraphics[width=0.85\linewidth]{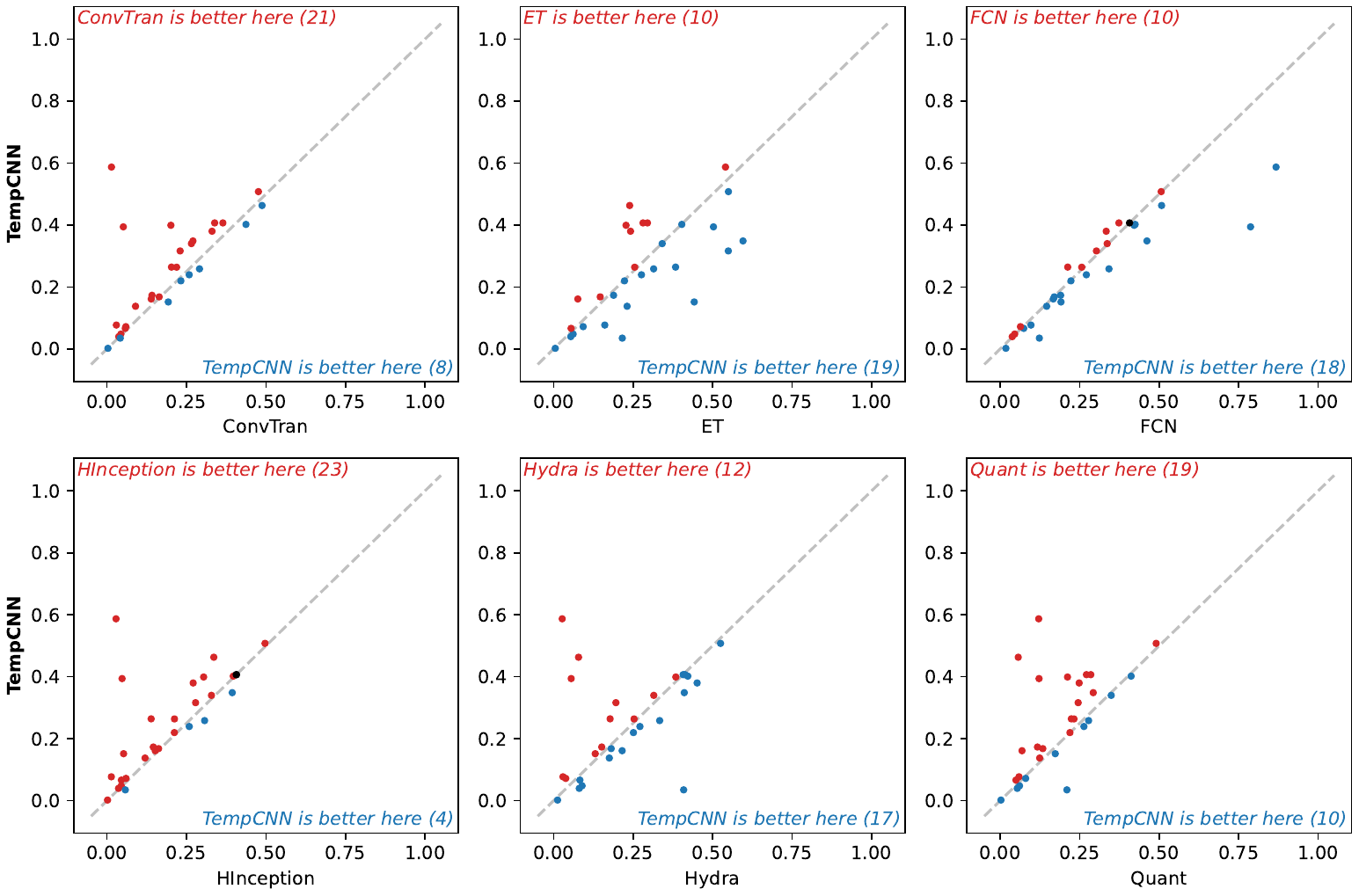}%
    \caption{Pairwise results ({\zo}) for TempCNN.}%
\end{figure}%

\clearpage


\subsection{Log Loss}

\begin{figure}[h]%
    \centering%
    \includegraphics[width=0.85\linewidth]{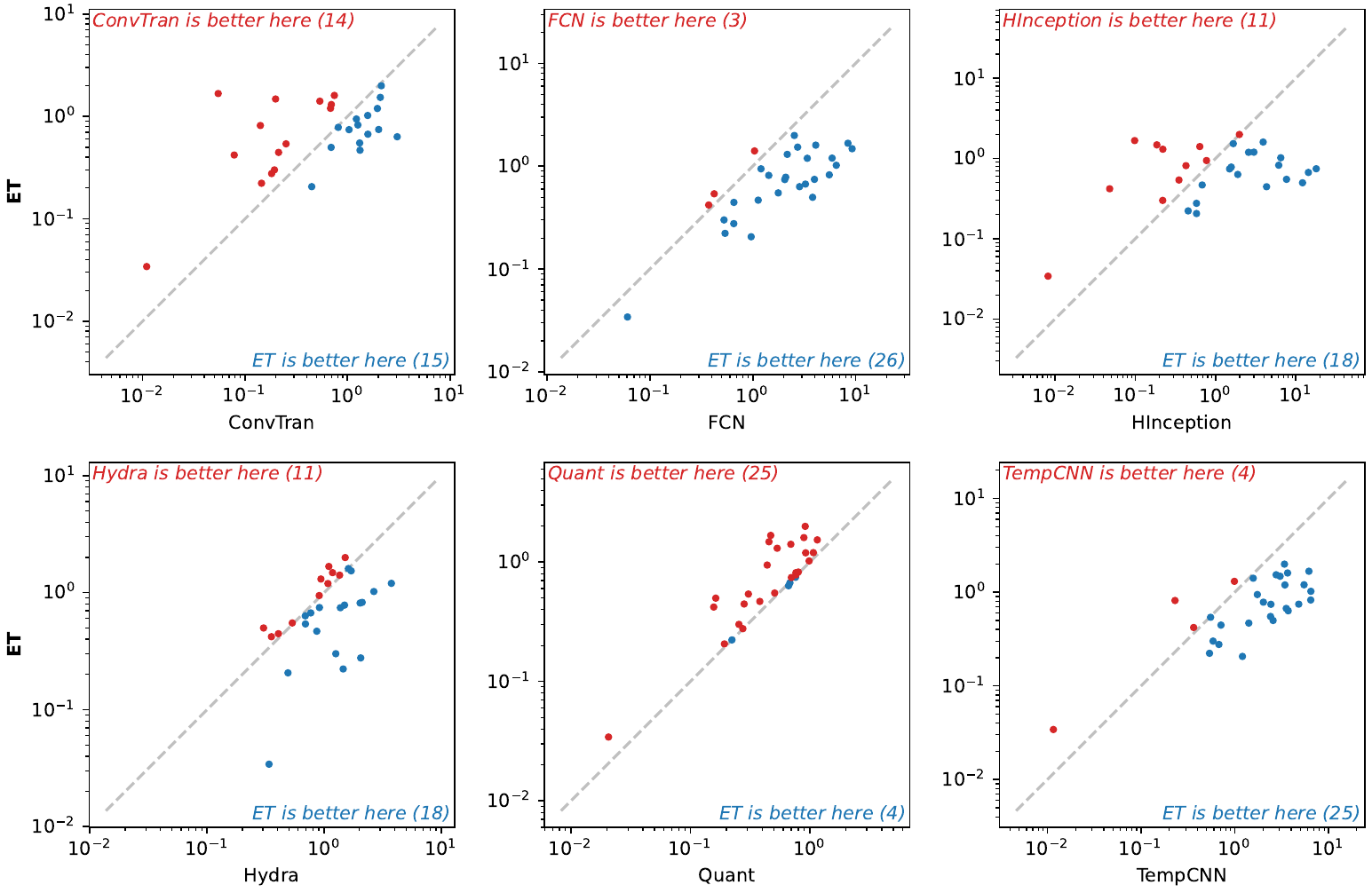}%
    \caption{Pairwise results (log-loss) for ET.}%
\end{figure}%

\begin{figure}[h]%
    \centering%
    \includegraphics[width=0.85\linewidth]{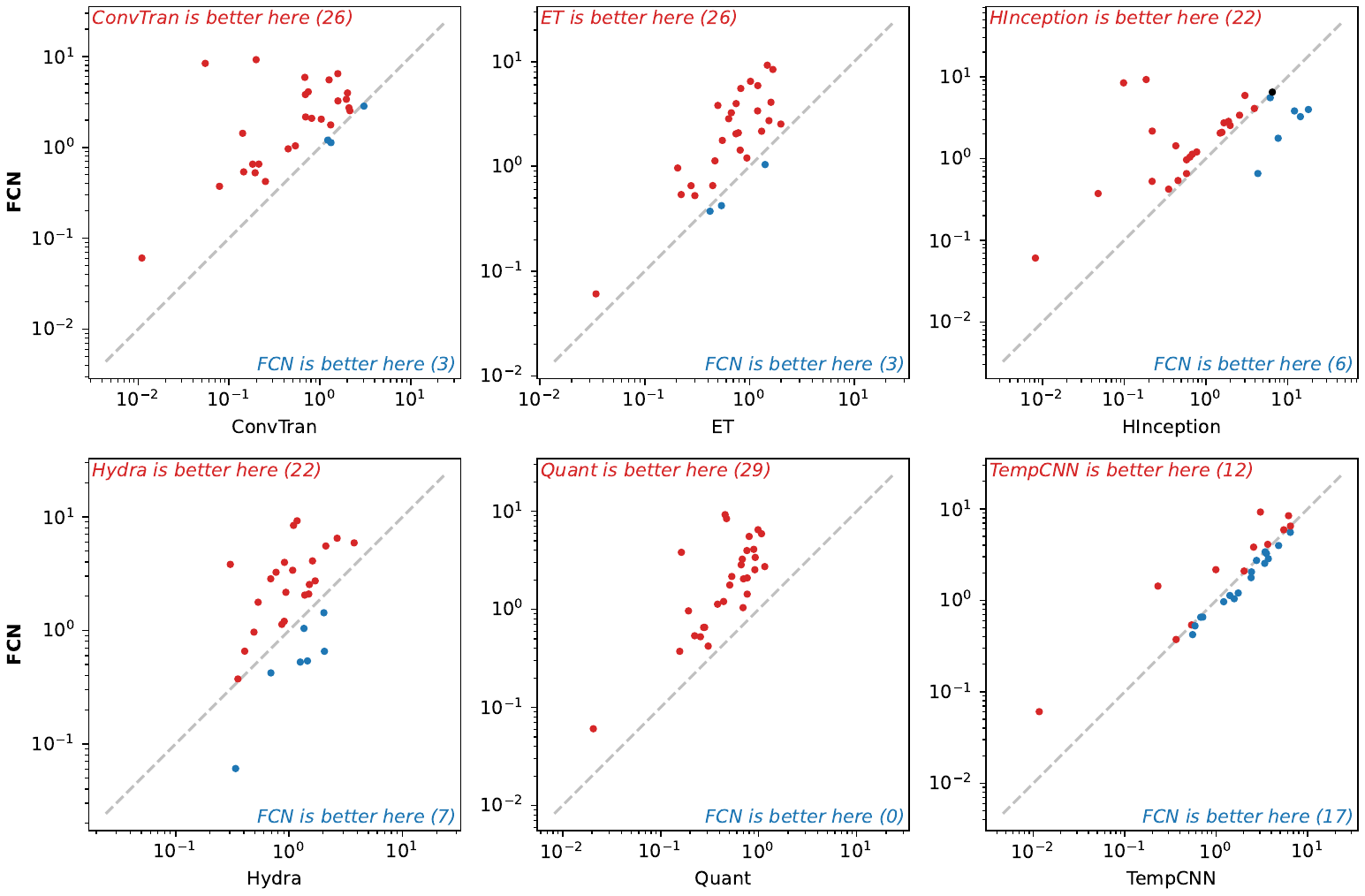}%
    \caption{Pairwise results (log-loss) for FCN.}%
\end{figure}%

\begin{figure}[h]%
    \centering%
    \includegraphics[width=0.85\linewidth]{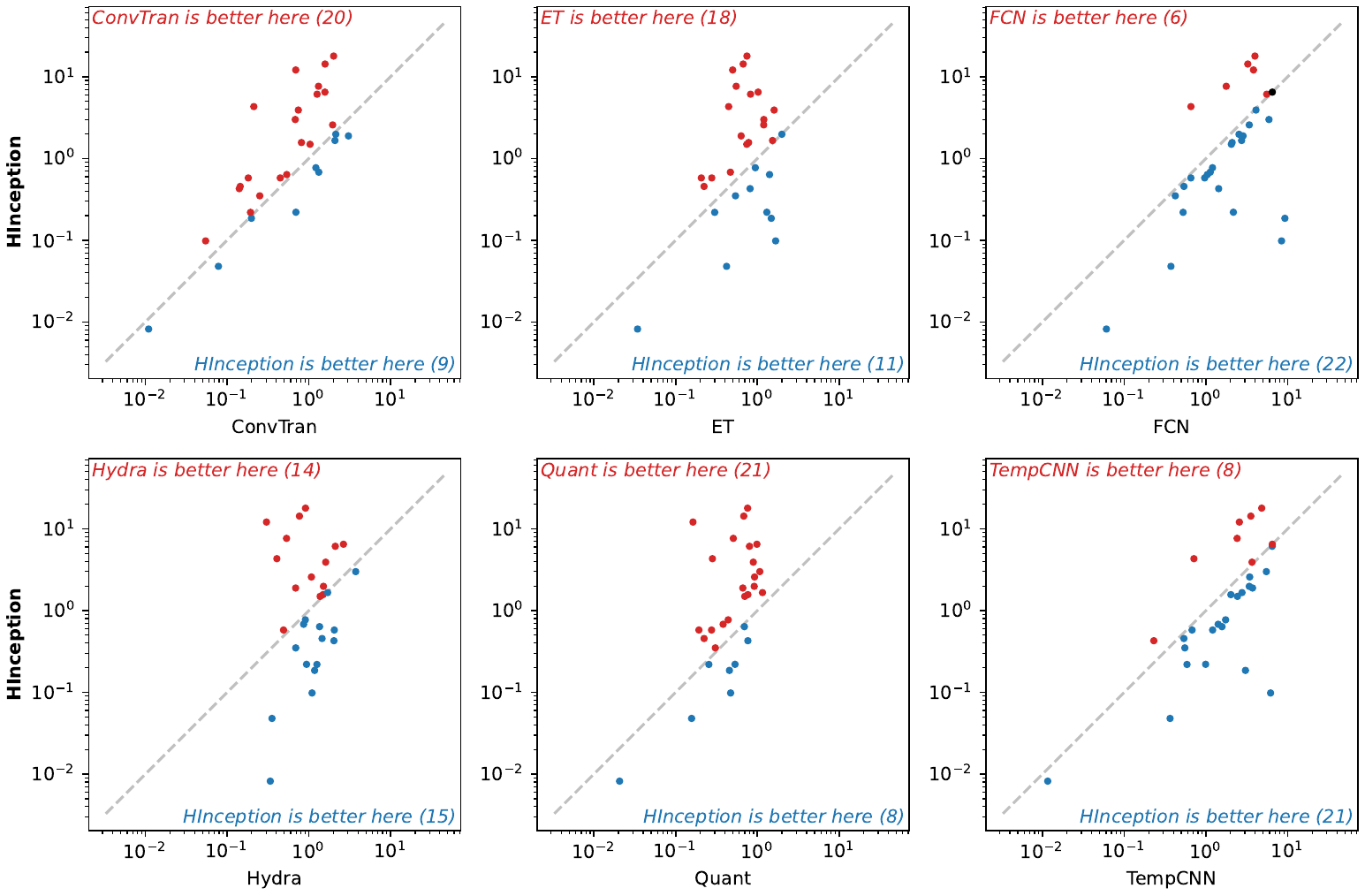}%
    \caption{Pairwise results (log-loss) for HInceptionTime.}%
\end{figure}%

\begin{figure}[h]%
    \centering%
    \includegraphics[width=0.85\linewidth]{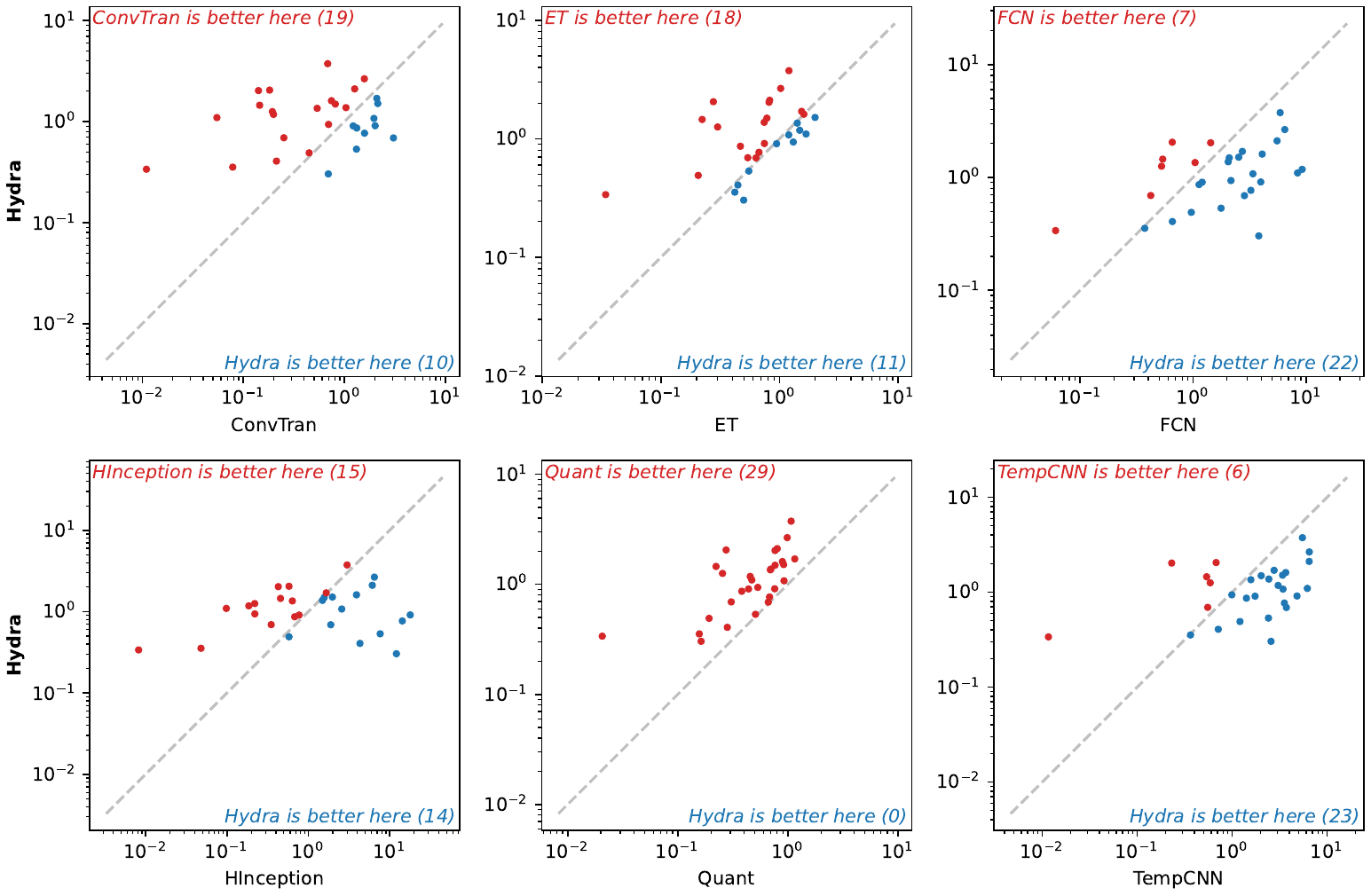}%
    \caption{Pairwise results (log-loss) for {\hydra}.}%
\end{figure}%

\begin{figure}[h]%
    \centering%
    \includegraphics[width=0.85\linewidth]{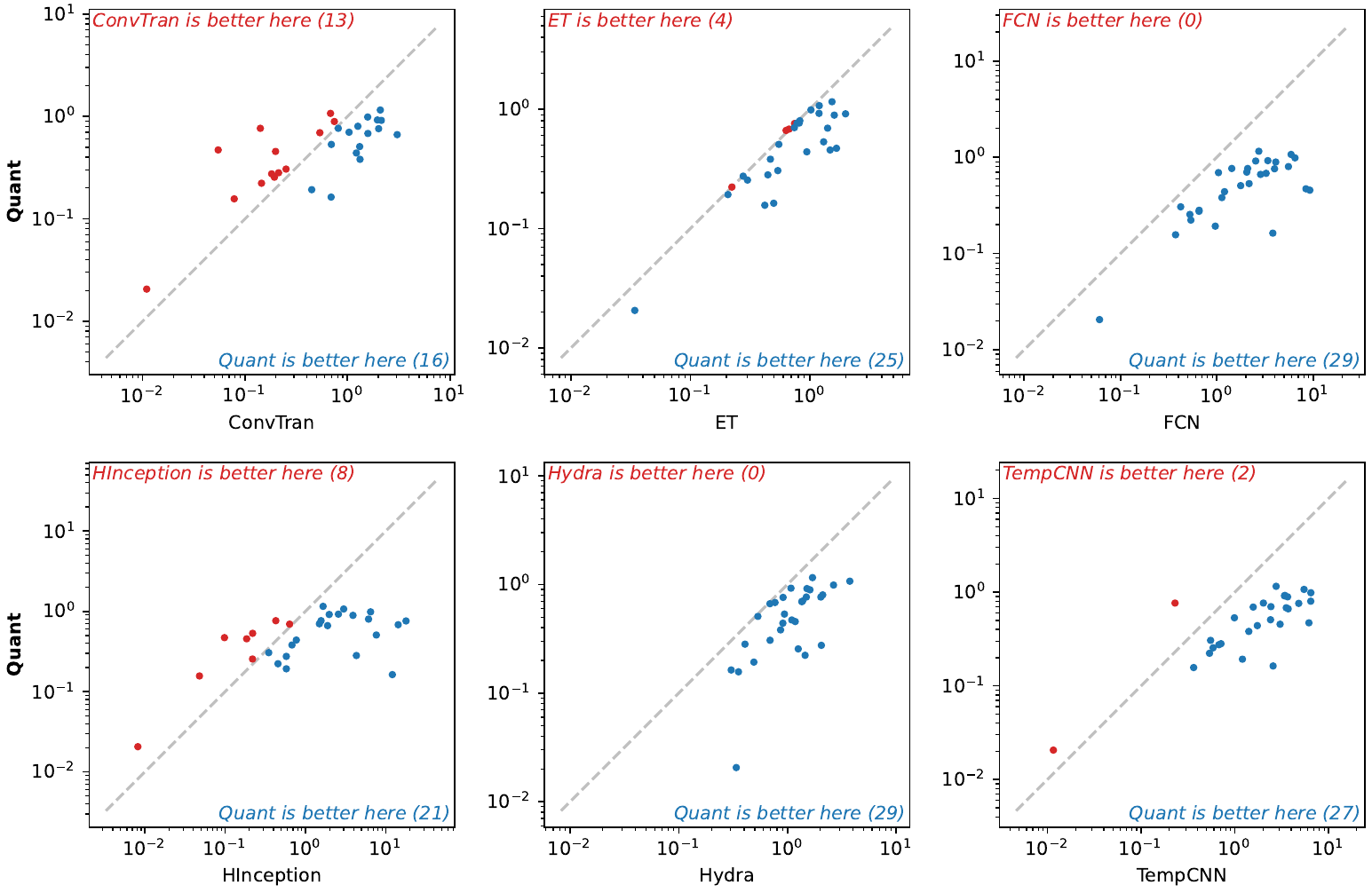}%
    \caption{Pairwise results (log-loss) for {\quant}.}%
\end{figure}%

\begin{figure}[h]%
    \centering%
    \includegraphics[width=0.85\linewidth]{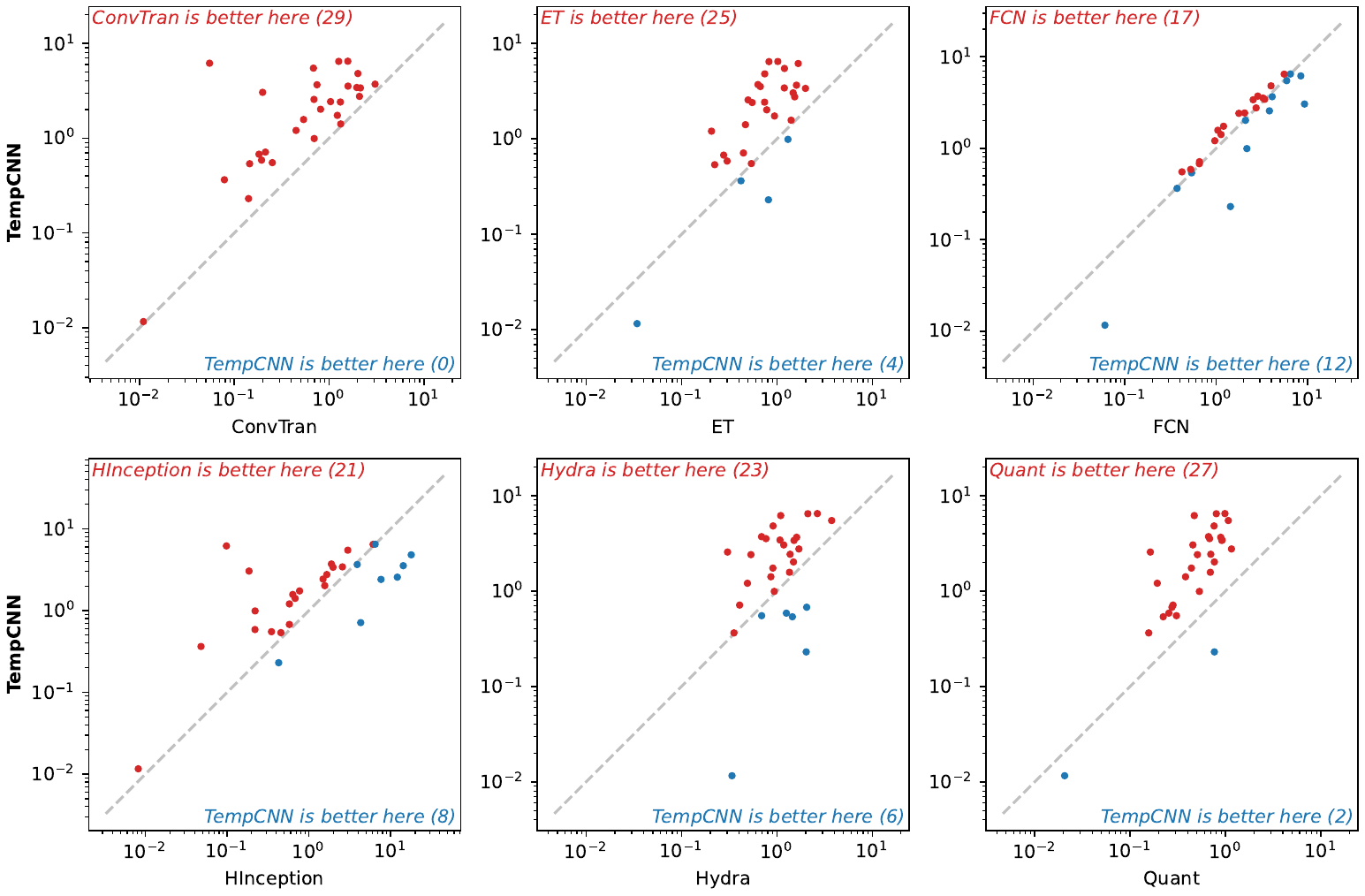}%
    \caption{Pairwise results (log-loss) for TempCNN.}%
\end{figure}%

\clearpage


\subsection{Training Time}

\begin{figure}[h]%
    \centering%
    \includegraphics[width=0.85\linewidth]{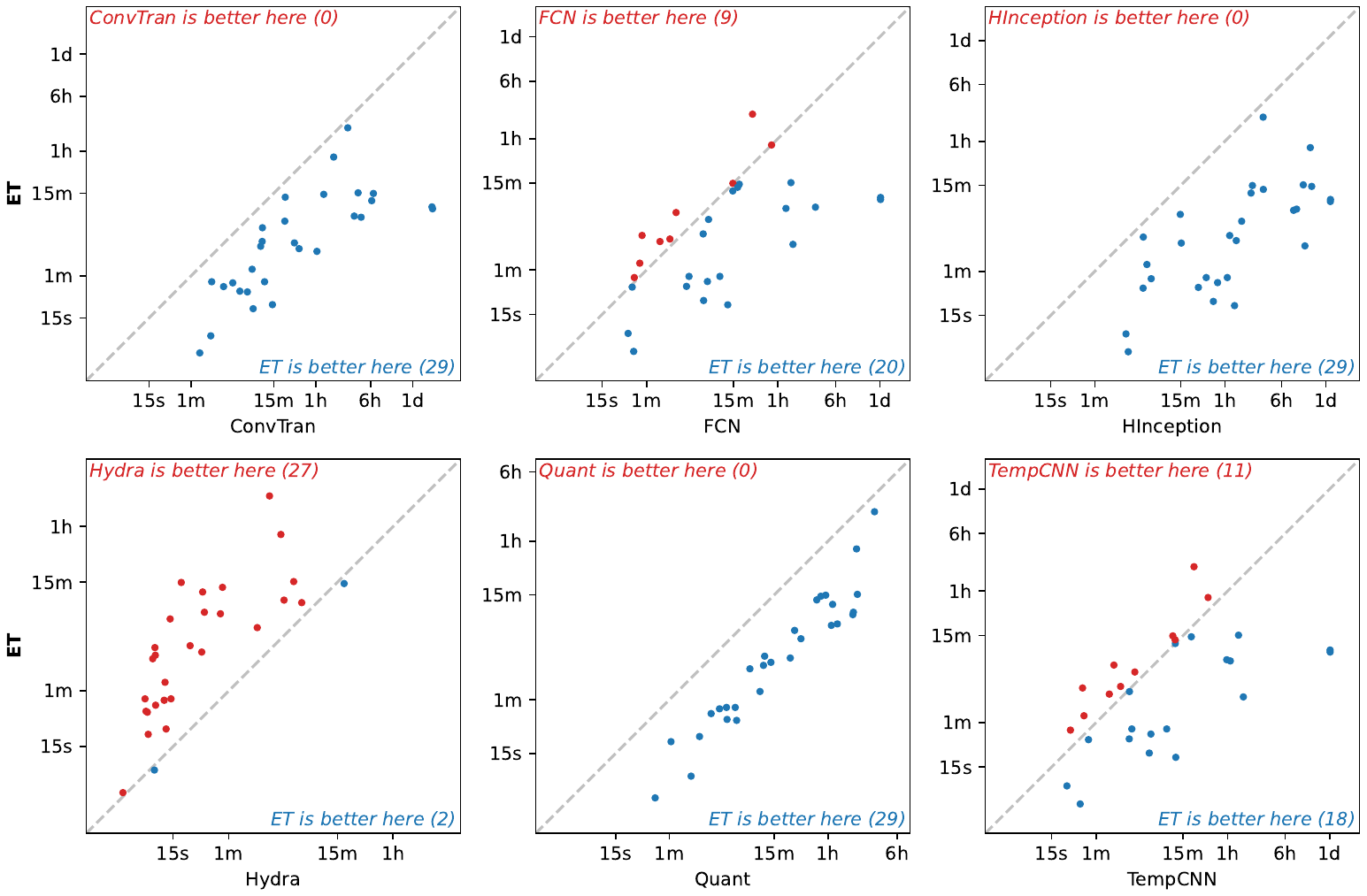}%
    \caption{Pairwise results (training time) for ET.}%
\end{figure}%

\begin{figure}[h]%
    \centering%
    \includegraphics[width=0.85\linewidth]{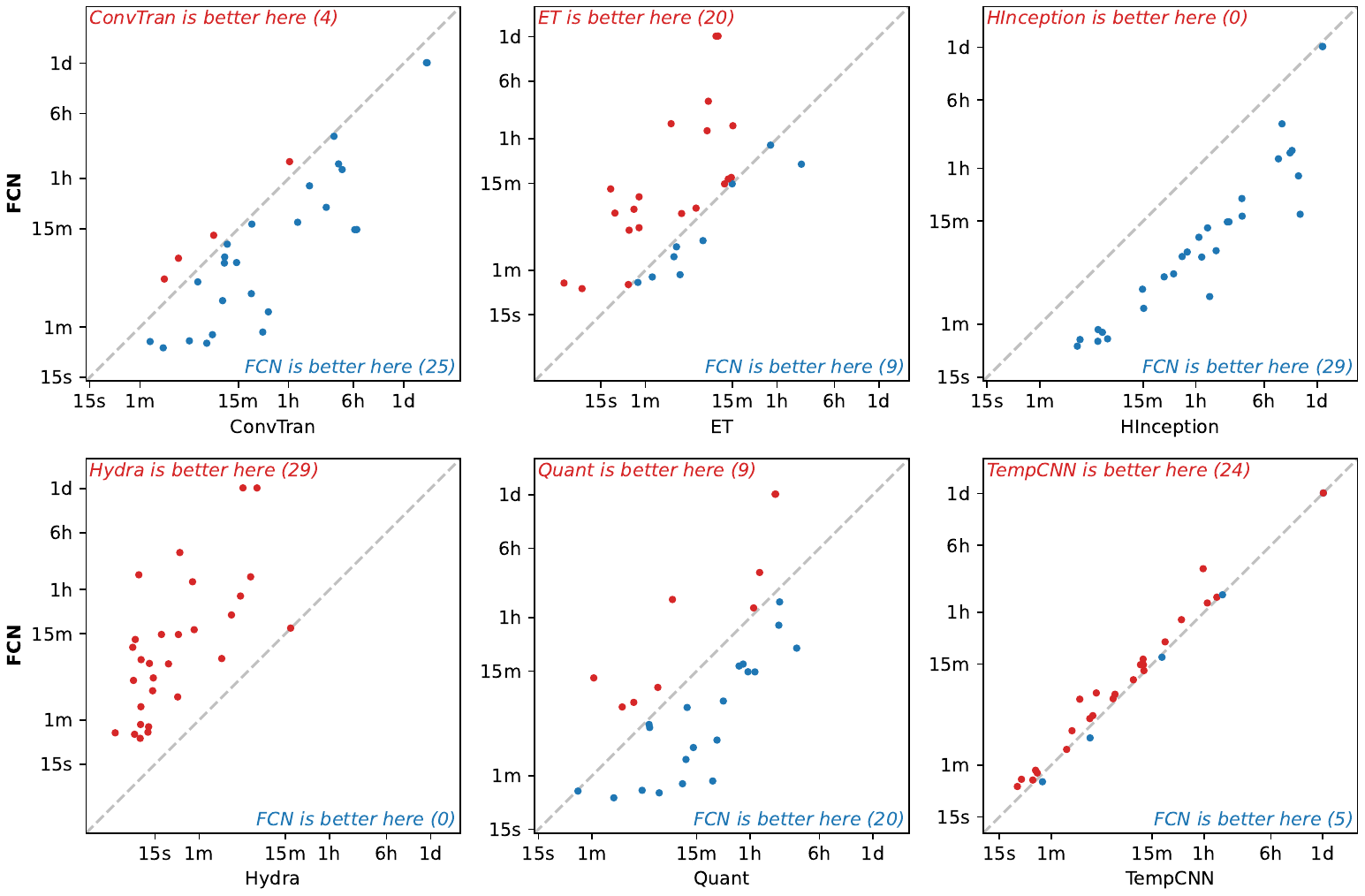}%
    \caption{Pairwise results (training time) for FCN.}%
\end{figure}%

\begin{figure}[h]%
    \centering%
    \includegraphics[width=0.85\linewidth]{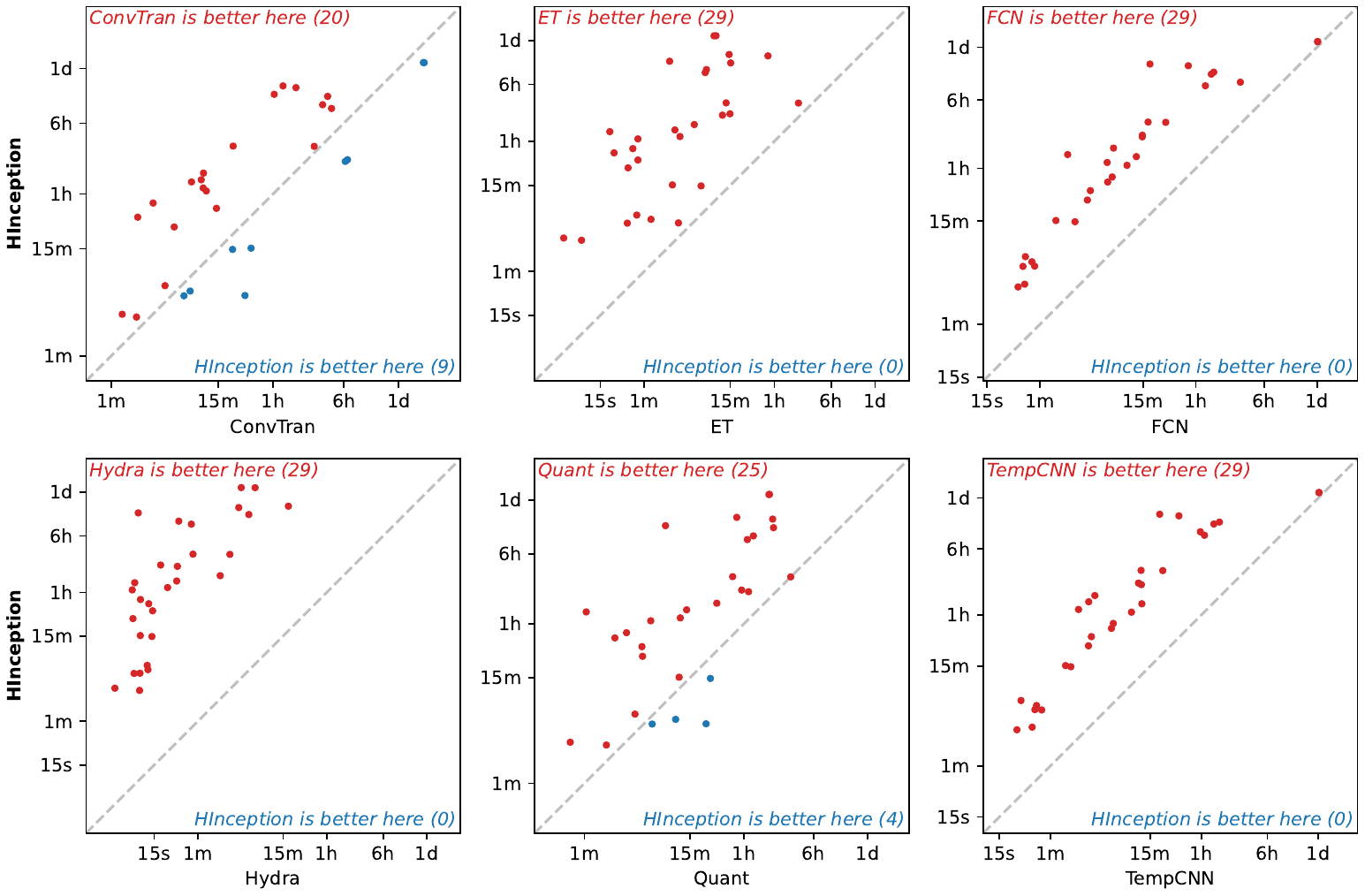}%
    \caption{Pairwise results (training time) for HInceptionTime.}%
\end{figure}%

\begin{figure}[h]%
    \centering%
    \includegraphics[width=0.85\linewidth]{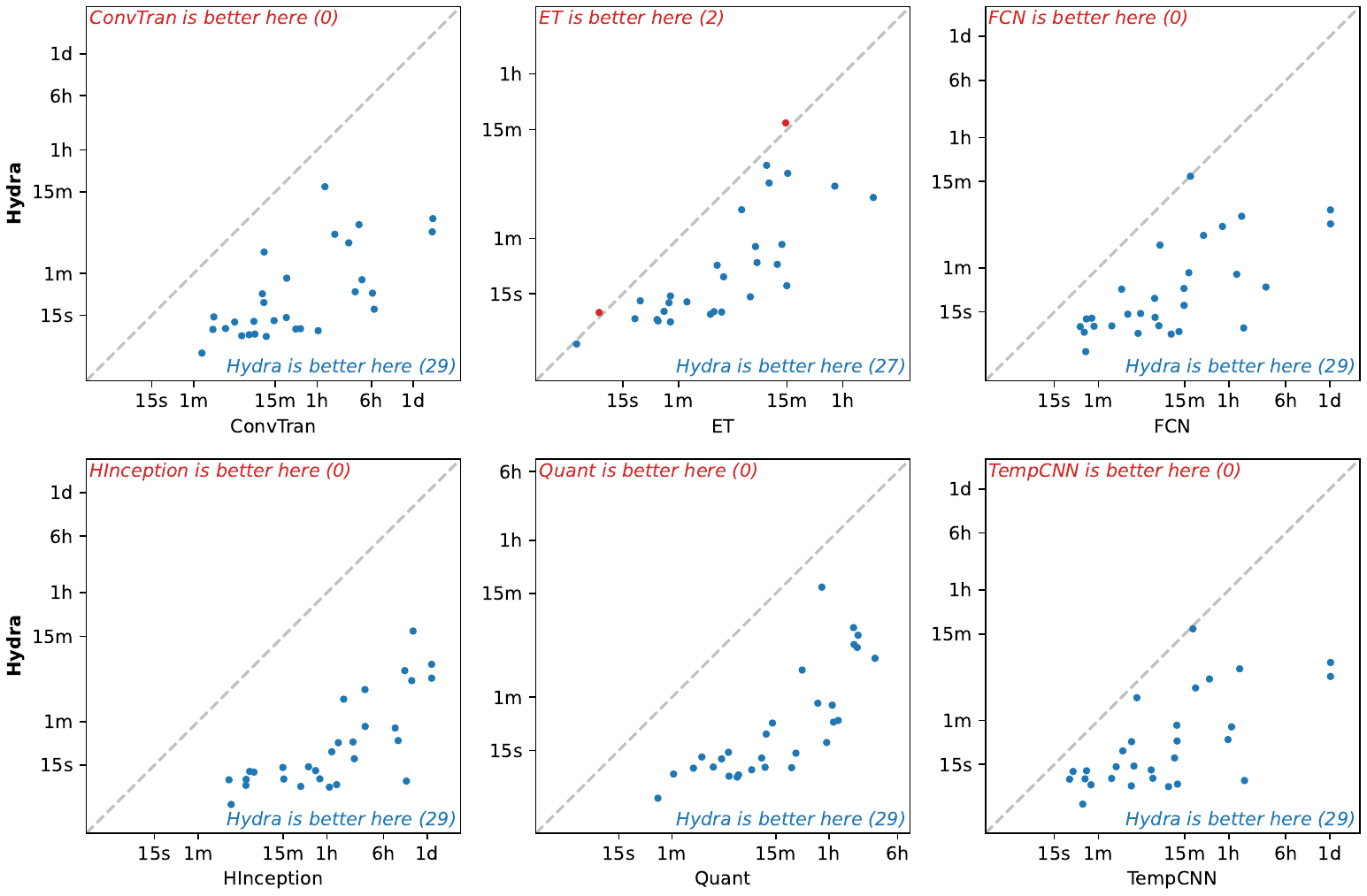}%
    \caption{Pairwise results (training time) for {\hydra}.}%
\end{figure}%

\begin{figure}[h]%
    \centering%
    \includegraphics[width=0.85\linewidth]{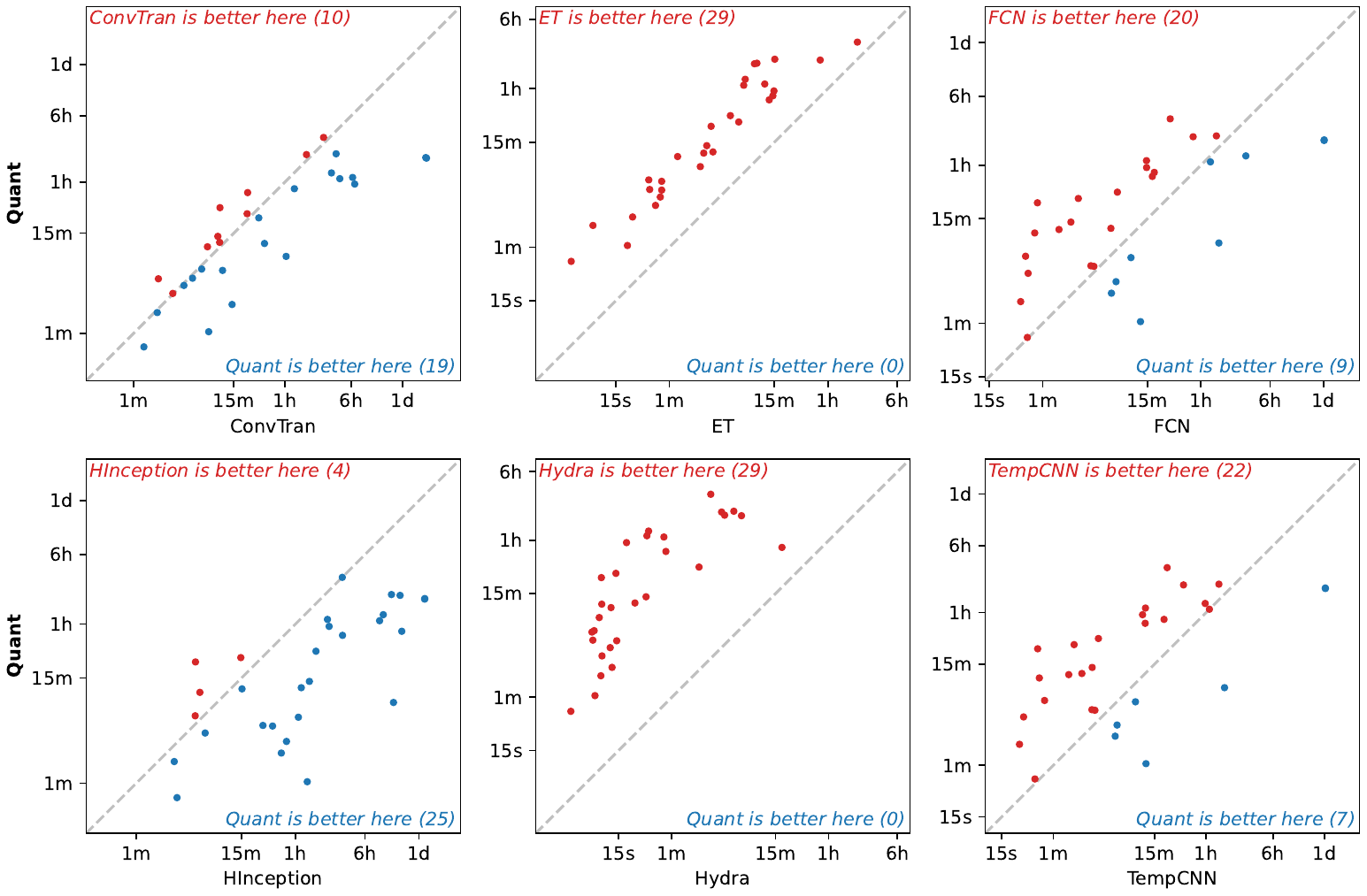}%
    \caption{Pairwise results (training time) for {\quant}.}%
\end{figure}%

\begin{figure}[h]%
    \centering%
    \includegraphics[width=0.85\linewidth]{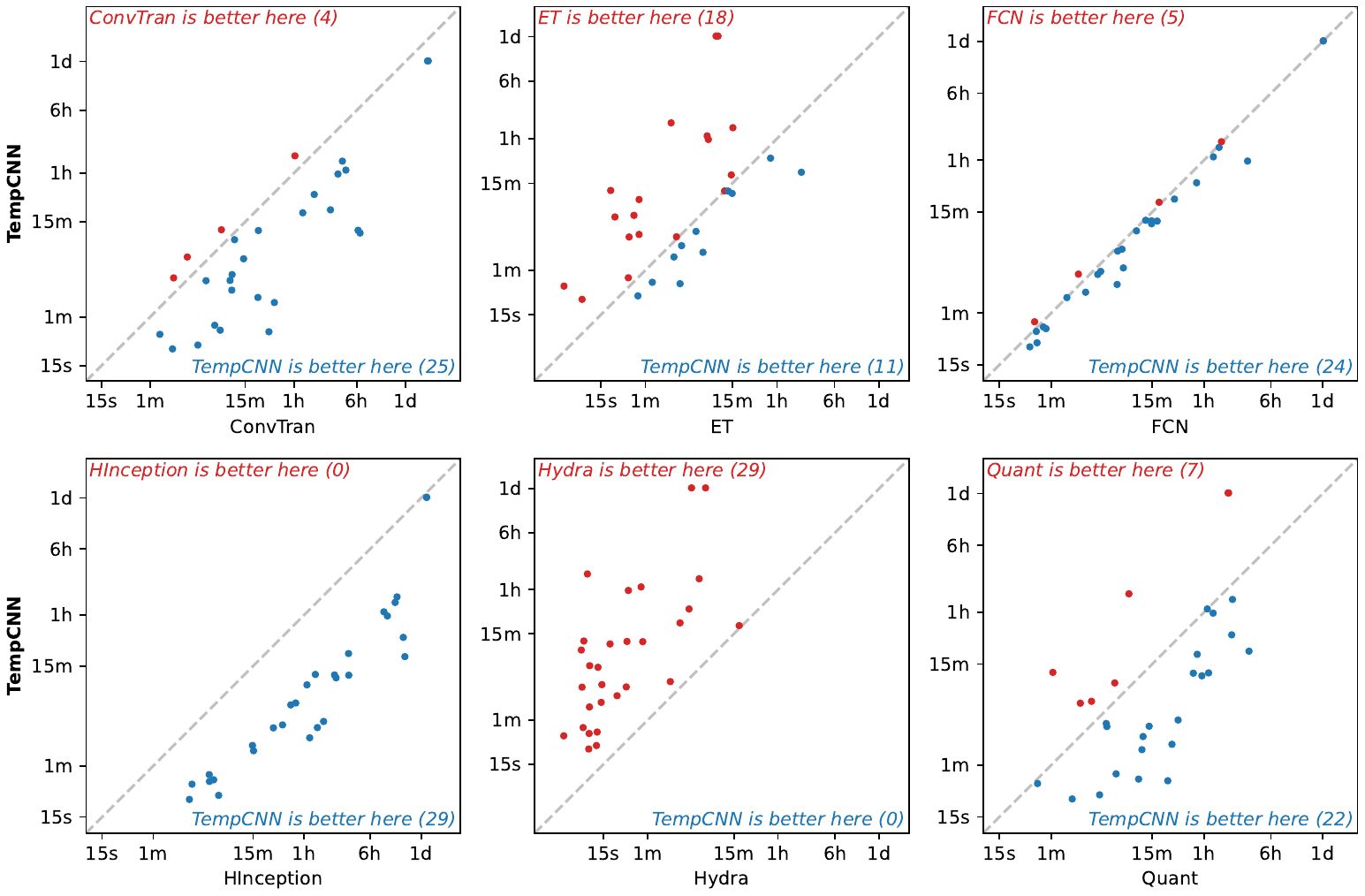}%
    \caption{Pairwise results (training time) for TempCNN.}%
\end{figure}%

\clearpage

\subsection{Learning Curves ({\hydra} and {\quant})} \label{sec-appendix-curves-hydra-quant}

\begin{figure}[h]
    \centering
    \includegraphics[width=0.70\linewidth]{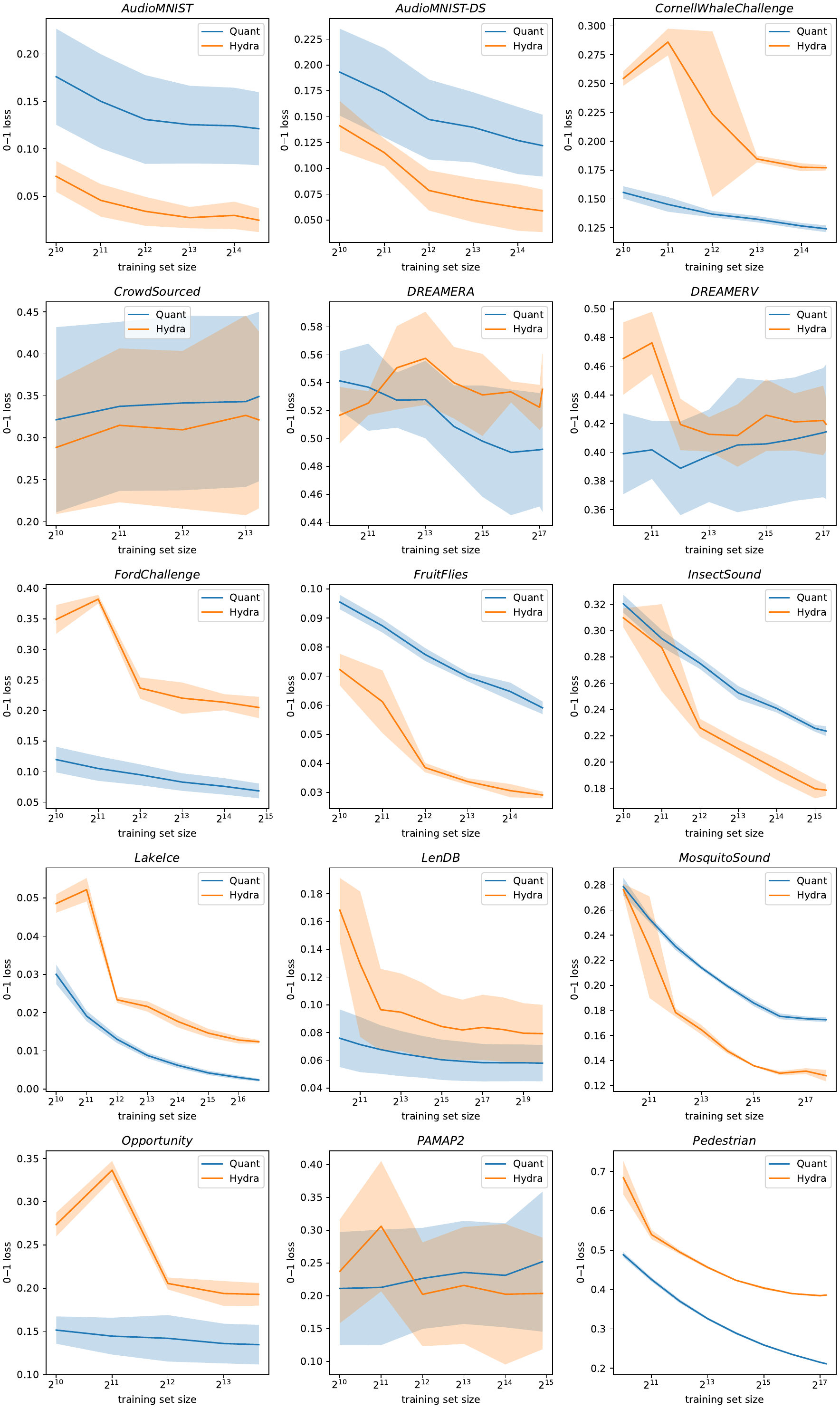}
    \caption{Learning curves (1 of 2).}%
\end{figure}%

\begin{figure}[h]%
    \centering%
    \includegraphics[width=0.70\linewidth]{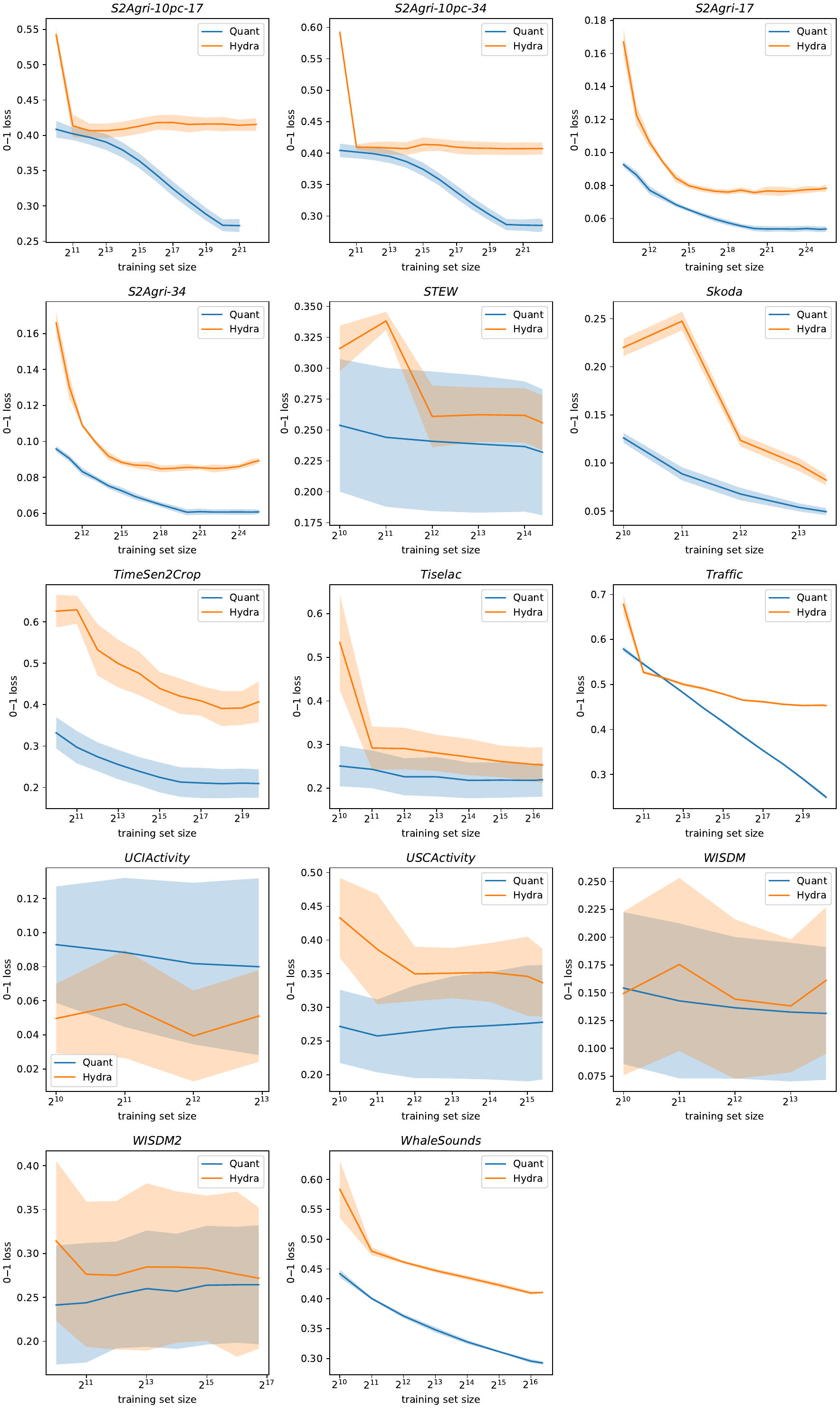}%
    \caption{Learning curves (2 of 2).}%
\end{figure}%

\clearpage

\section{Bias--Variance Learning Curves Detail} \label{sec-appendix-bv}

The low variance model comprises a single layer of 100 fixed, random convolutional kernels. The low bias model comprises two layers of 100 learned convolutional kernels. Both models use ReLU activations, with global average pooling followed by batch normalisation and a learned linear layer. The input time series are normalised by subtracting the mean and dividing by the standard deviation (per series). Both models were trained for 500 epochs for each dataset size using the Adam optimizer, with no explicit regularisation, using an initial learning rate of $10^{-2}$ reduced by a factor of $10$ after each 100 epochs.

\end{document}